\documentclass[12pt,onecolumn,draftclsnofoot]{IEEEtran}
\usepackage{cite}

\ifCLASSINFOpdf
  \usepackage[pdftex]{graphicx}
	\usepackage{caption, adjustbox,setspace}
  \usepackage{subcaption}
	\usepackage{subfloat}
\else
\fi

\usepackage{enumerate, booktabs, amsfonts, pifont, paralist}

\usepackage{tikz}
\usetikzlibrary{shapes,arrows}
\usepackage{array}
\usepackage{amsmath}
\usepackage{mdframed}
\usepackage{xcolor}
\usepackage{varwidth}
\usepackage{algpseudocode}
\usepackage{multirow}

\usepackage{float}
\usepackage{verbatim}

\begin{document}
%
\title{\mbox{Full Reference Objective Quality Assessment}\\for Reconstructed Background Images}
%
%
%

\author{Aditee~Shrotre,~\IEEEmembership{Student Member,~IEEE,}
        and~Lina~J~Karam,~\IEEEmembership{Fellow,~IEEE}
\thanks{ Copyright (c) 2017 IEEE. Personal use of this material is permitted. However, permission to use this material for any other purposes must be obtained from the IEEE by sending a request to pubs-permissions@ieee.org. \newline The authors are with the School of Electrical, Computer and Energy Engineering, Arizona State University, Tempe, AZ, 85287 E-mail: \{ashrotre,karam\}@asu.edu.}
\thanks
}

%
%

\markboth{IEEE TRANSACTIONS ON IMAGE PROCESSING,~2017}%
{Shrotre \MakeLowercase{\textit{et al.}}:  Database and Evaluation technique for Background reconstruction }
%



\maketitle

\vspace{-10ex}
\begin{abstract}
With an increased interest in applications that require a clean background image, such as video surveillance, object tracking, street view imaging and location-based services on web-based maps, multiple algorithms have been developed to reconstruct a background image from cluttered scenes. 
Traditionally, statistical measures and existing image quality techniques have been applied for evaluating the quality of the reconstructed background images. 
Though these quality assessment methods have been widely used in the past, their performance in evaluating the perceived quality of the reconstructed background image has not been verified. 
In this work, we discuss the shortcomings in existing metrics and propose a full reference Reconstructed Background image Quality Index (RBQI) that combines color and structural information at multiple scales using a probability summation model to predict the perceived quality in the reconstructed background image given a reference image. 
To compare the performance of the proposed quality index with existing image quality assessment measures, we construct two different datasets consisting of reconstructed background images and corresponding subjective scores. The quality assessment measures are evaluated by correlating their objective scores with human subjective ratings. The correlation results show that the proposed RBQI outperforms all the existing approaches.
Additionally, the constructed datasets and the corresponding subjective scores provide a benchmark to evaluate the performance of future metrics that are developed to evaluate the perceived quality of reconstructed background images.
\end{abstract}


%
\IEEEpeerreviewmaketitle

\section{Introduction}
%
%
%
%
\IEEEPARstart{A}{} clean background image has great significance in multiple applications. It can be used for video surveillance \cite{Surveillance}, activity recognition \cite{Activity}, object detection and tracking \cite{ObjectDetection}, \cite{Tracking}, street view imaging and location-based services on web-based maps \cite{StreetView, VirtualMap}, and texturing 3D models obtained from multiple photographs or videos \cite{3DModeling}. But acquiring a clean photograph of a scene is seldom possible. There are always some unwanted objects occluding the background of interest. The technique of acquiring a clean background image by removing the occlusions using frames from a video or multiple views of a scene, is known as background reconstruction or background initialization. Many algorithms have been proposed for initializing the background images from videos, for example, \cite{SOBS, SpatioTemp, bkgEstimator, BTS, MRF, MultiLayer, PBI};  and also from multiple images such as \cite{Cormac, Photomontage, CnGrad}. 

Background initialization or reconstruction is crippled by multiple challenges. The pseudo-stationary background (e.g., waving trees, waves in water, etc.) poses additional challenges in separating the moving foreground objects from the relatively stationary background pixels. The illumination conditions can vary across the images thus changing the global characteristics of each image. The illumination changes cause local phenomena such as shadows, reflections and shading, which change the local characteristics of the background across the images or frames in a video. Finally, the removal of ‘foreground’ objects from the scene creates holes in the background that need to be filled in with pixels that maintain the continuity of the background texture and structures in the recovered image.
{\parskip 0pt
Thus the background reconstruction algorithms can be characterized by two main tasks: 
\begin{inparaenum}
	\item foreground detection, in which the foreground is separated from the background by classifying pixels as foreground or background;
	\item background recovery, in which the holes formed due to foreground removal are filled.
\end{inparaenum} 
}

{\parskip 0pt 
The performance of a background extraction algorithm depends on two factors: \begin{inparaenum} 
	\item its ability to detect the foreground objects in the scene and completely eliminate them;  and
	\item the perceived quality of the reconstructed background image.
\end{inparaenum} 
Traditional statistical techniques such as Peak Signal to Noise Ratio (PSNR), Average Gray-level Error (AGE), total number of error pixels (EPs), percentage of EPs (pEP), number of Clustered Error Pixels (CEPs) and percentage of CEPs (pCEPs)~\cite{BkgInitDataset} quantify the performance of the algorithm in its ability to remove foreground objects from a scene to a certain extent, but they do not give an indication of the perceived quality of the generated background image. On the other hand, the existing Image Quality Assessment (IQA) techniques such as Multi-scale Similarity metric (MS-SSIM)~\cite{MSSIM} and Color image Quality Measures (CQM)~\cite{CQM} used by the authors in~\cite{BI_Taxonomy} to compare different background reconstruction algorithms are not designed to identify any residual foreground objects in the scene. 
Lack of a quality metric that can reliably assess the performance of background reconstruction algorithms by quantifying both aspects of a reconstructed background image motivated the development of the proposed Reconstructed Background visual Quality Index (RBQI). RBQI uses the contrast, structure and color information to determine the presence of any residual foreground objects in the reconstructed background image as compared to the reference background image and to detect any unnaturalness introduced by the reconstruction algorithm that affects the perceived quality of the reconstructed background image. 

This paper also presents two datasets that are constructed to assess the performance of the proposed as well as popular existing objective quality assessment methods in predicting the perceived visual quality of the reconstructed background images. The datasets consist of reconstructed background images generated using different background reconstruction algorithms in the literature along with the corresponding subjective ratings. 
Some of the existing datasets such as video surveillance datasets (Wallflower \cite{Wallflower}, I2R \cite{I2R}), background subtraction datasets (UCSD \cite{UCSD}, CMU \cite{CMU}) and object tracking evaluation dataset (``Performance Evaluation of Tracking and Surveillance (PETS)") are not suited for this application as they do not provide reconstructed background images but just the foreground masks as ground-truth.
The more recent database ``Scene Background Modeling Net" (SBMNet) \cite{SBMnet} is targeted at comparing the performance of the background initialization algorithms but it does not provide any subjective ratings for the reconstructed background images. Hence the SBMNet database \cite{SBMnet} is not suited for benchmarking the performance of objective background visual quality assessment.
The datasets proposed in this work are the first and currently the only datasets that can be used for benchmarking existing and future metrics developed to assess the quality of reconstructed background images.

{\parskip 0pt 
The rest of the paper is organized as follows. In Section~\ref{sec:survey} we highlight the limitations of existing popular assessment methods~\cite{QoMex}. 
We introduce the new benchmarking datasets in Section~\ref{sec:SQA} along with the details of the subjective tests. 
In Section~\ref{sec:Proposed}, we propose a new index that makes use of a probability summation model to combine structure and color characteristics at multiples scales for quantifying the perceived quality in reconstructed background images. 
Performance evaluation results for the existing and proposed objective visual quality assessment methods are presented in Section~\ref{sec:results} for reconstructed background images. 
Finally, we conclude the paper in Section~\ref{sec:conclusions} and also provide directions for future research.
}

\section{Existing Full Reference Background Quality Assessment Techniques and their limitations}
\label{sec:survey}
Existing background reconstruction quality metrics can be classified into two categories: statistical and image quality assessment (IQA) techniques, depending on the type of features used for measuring the similarity between the reconstructed background image and reference background image. 

\subsection{Statistical Techniques}
Statistical techniques use intensity values at co-located pixels in the reference and reconstructed background images to measure the similarity. Popular statistical techniques \cite{BkgInitDataset} that have been traditionally used for judging the performance of background initialization algorithms are briefly explained here.

(i) Average Gray-level Error ($AGE$): AGE is calculated as the absolute difference between the gray levels of the co-located pixels in the reference and reconstructed background image.

(ii) Error Pixels ($EP$):  $EP$ gives the total number of error pixels. A pixel is classified as an error pixel if the absolute difference between the corresponding pixels in the reference and reconstructed background images is greater than an empirically selected threshold $\tau$.

(iii) Percentage Error Pixels ($pEP$): Percentage of the error pixels, calculated as ${EP}/{N}$, where $N$ is the total number of pixels in the image.

(iv) Clustered Error Pixels ($CEP$): $CEP$ gives the total number of clustered error pixels. A clustered error pixel is defined as the error pixel whose 4 connected pixels are also classified as error pixels.

(v) Percentage Clustered Error Pixels ($pCEP$): Percentage of the clustered error pixels, calculated as ${CEP}/{N}$, where $N$ is the total number of pixels in the image.

Though these techniques have been used to judge the quality of the reconstructed background images, their performance has not been previously evaluated. As we show in Section~\ref{sec:results} and as noted by the authors in \cite{QoMex}, the statistical techniques were found to not correlate well with the subjective quality scores.

\subsection{Image Quality Assessment}
The existing Full Reference Image Quality Assessment (FR-IQA) techniques use perceptually inspired features for measuring the similarity between two images. Though these techniques have been shown to work reasonably well while assessing images affected by distortions such as blur, compression artifacts and noise, these techniques have not been designed for assessing the quality of reconstructed background images. 
In \cite{BI_Taxonomy} popular FR-IQA techniques including Peak Signal to Noise ratio (PSNR), Multi-scale Similarity metric (MS-SSIM)~\cite{MSSIM} and Color image Quality Measure (CQM)~\cite{CQM}, were adopted for objectively comparing the performance of the different background reconstruction algorithms; however, no performance evaluation was carried out to support the choice of these techniques.
Other popular IQA techniques include Structural Similarity Index (SSIM)~\cite{SSIM}, visual signal-to-noise ratio (VSNR) \cite{VSNR}, visual information fidelity (VIF) \cite{VIF}, pixel-based VIF (VIFP) \cite{VIF}, universal quality index (UQI) \cite{UQI}, image fidelity criterion (IFC) \cite{IFC}, noise quality measure (NQM) \cite{NQM}, weighted signal-to-noise ratio (WSNR) \cite{WSNR}, feature similarity index (FSIM) \cite{FSIM}, FSIM with color (FSIMc) \cite{FSIM}, spectral residual based similarity (SR-SIM) \cite{SRSIM} and saliency-based SSIM (SalSSIM) \cite{SalSSIM}. The suitability of these techniques for evaluating the quality of reconstructed background images remains unexplored.

As the first contribution of this paper we present two benchmarking datasets that can be used for comparing the performance of different techniques in objectively assessing the perceived quality of the reconstructed background images. These datasets contain reconstructed background images along with their subjective ratings, details of which are discussed in Section~\ref{sec:database}. 
When the statistical and IQA techniques were tested on these datasets, none of the techniques were found to correlate well with the subjective scores as discussed in Section~\ref{sec:results}. This motivated our second contribution, the objective Reconstructed Background Quality Index (RBQI) that is shown to outperform all the existing techniques in assessing the perceived visual quality of reconstructed background images.

\section{Subjective Quality Assessment of Reconstructed Background Images} 
\label{sec:SQA}
\subsection{Databases}
\label{sec:database}
In this section we present two different datasets constructed as part of this work to serve as benchmarks for comparing existing and future techniques developed for assessing the quality of reconstructed background images. The images and subjective experiments for both datasets are described in the subsequent subsections.

Each dataset contains the original sequence of images or videos that are used as inputs to the different reconstruction algorithms, the background images reconstructed by the different algorithms and the corresponding subjective scores. 

\subsubsection{Reconstructed Background Quality (ReBaQ) Dataset}
\label{sec:dataset1}

\begin{figure}[tp]
  \centering
	\begin{subfigure}[t]{\textwidth}
		\begin{subfigure}[t]{0.24\textwidth}	
			\captionsetup{labelformat=empty, format=default}
			{\includegraphics[width=4cm]{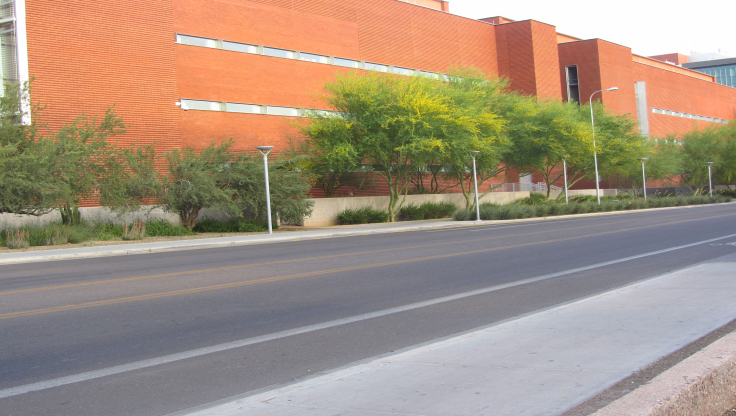}}\hfill%
			\subcaption{Street \\ (Outdoor Scene)}
		\end{subfigure}
		\begin{subfigure}[t]{0.24\textwidth}
			\captionsetup{labelformat=empty, format=default}
			{\includegraphics[width=4cm]{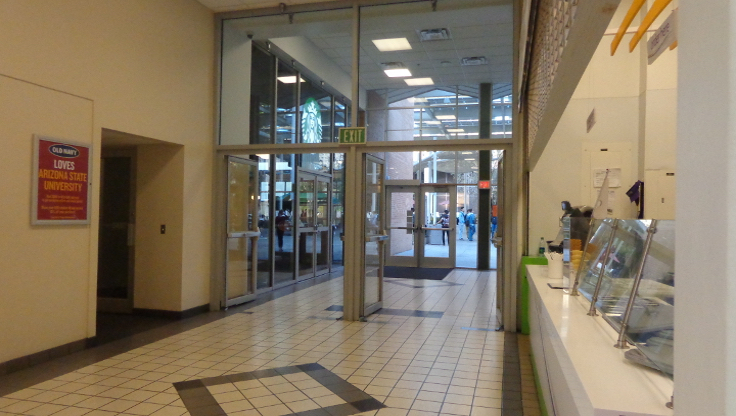}}\hfill%
			\subcaption{Hall \\ (Indoor Scene)}
		\end{subfigure}
		\begin{subfigure}[t]{0.24\textwidth}
			\captionsetup{labelformat=empty, format=default}
			{\includegraphics[width=4cm]{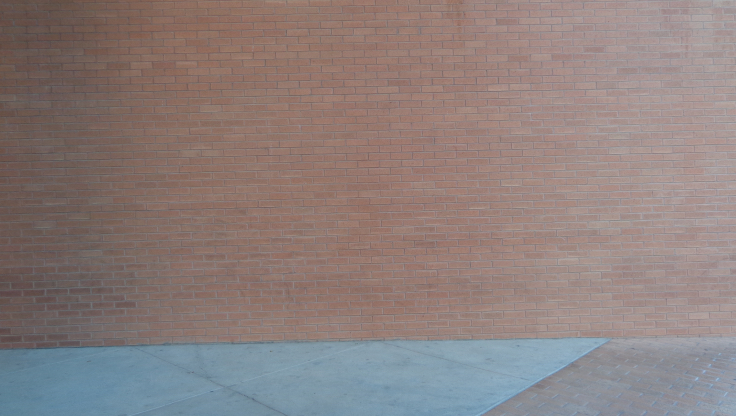}}\hfill%
			\subcaption{Wall \\ (Textured background)}
		\end{subfigure}
		\begin{subfigure}[t]{0.24\textwidth}
			\captionsetup{labelformat=empty, format=default}
			{\includegraphics[width=4cm]{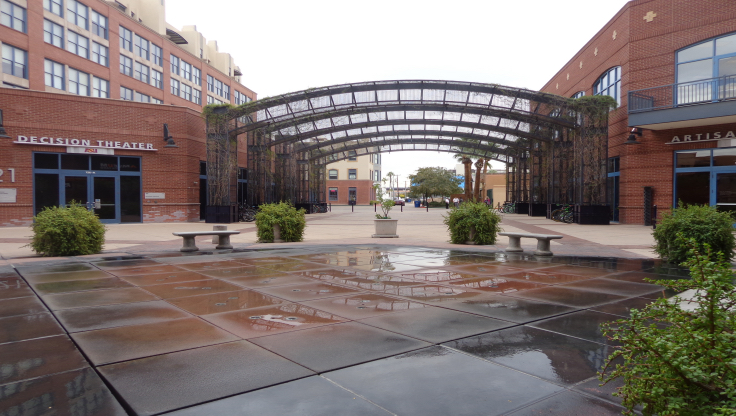}}\hfill%
			\subcaption{\centering WetFloor \newline(Water as low-contrast foreground)}
		\end{subfigure}
	\setcounter{subfigure}{0}
	\caption{Scenes with static backgrounds from the ReBaQ dataset.}
	\label{fig:staticReBaQ}
	\end{subfigure} 
\par\bigskip
	\begin{subfigure}[t]{\textwidth}
		\begin{subfigure}[t]{0.24\textwidth}	
			\captionsetup{labelformat=empty, format=default}
			{\includegraphics[width=4cm]{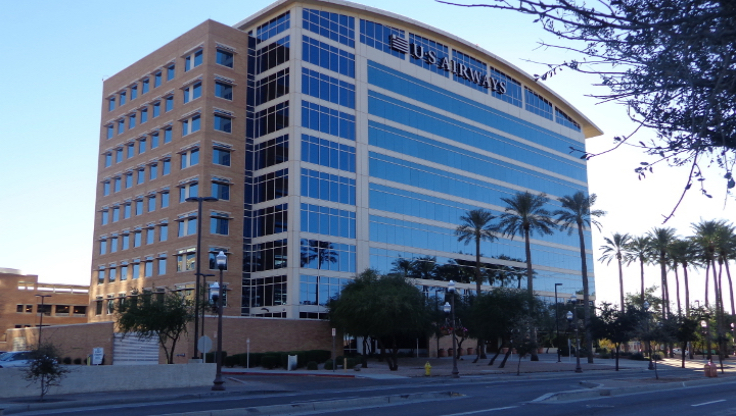}}\hfill%
			\subcaption{Building \\ (Reflective)}
		\end{subfigure}
		\begin{subfigure}[t]{0.24\textwidth}
			\captionsetup{labelformat=empty, format=default}
			{\includegraphics[width=4cm]{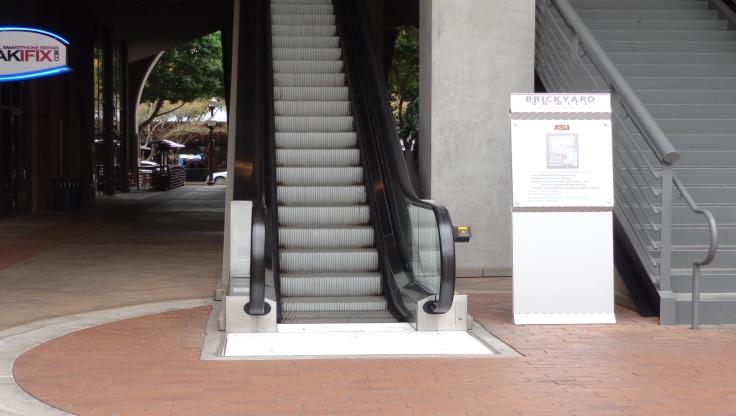}}\hfill%
			\subcaption{Escalator \\ (Large Motion)}
		\end{subfigure}
		\begin{subfigure}[t]{0.24\textwidth}
			\captionsetup{labelformat=empty, format=default}
			{\includegraphics[width=4cm]{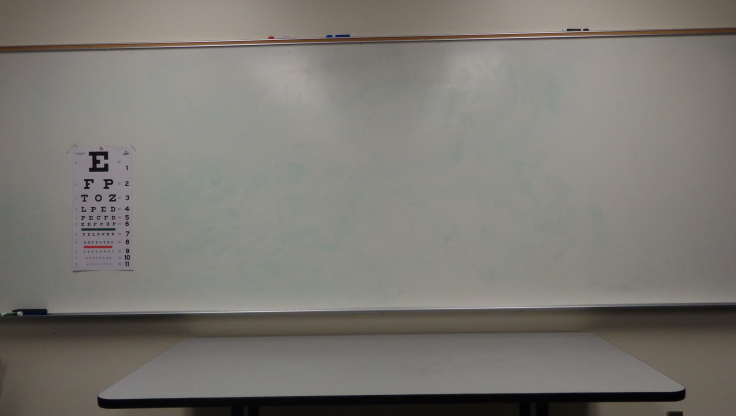}}\hfill%
			\subcaption{\centering llumination \newline (Illumination Variations)}
		\end{subfigure}
		\begin{subfigure}[t]{0.24\textwidth}
			\captionsetup{labelformat=empty, format=default}
			{\includegraphics[width=4cm]{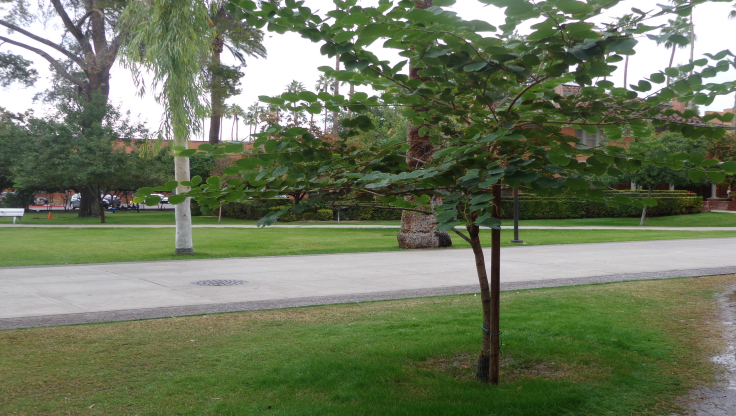}}\hfill%
			\subcaption{Park \\ (Small Motion)}
		\end{subfigure}
	\setcounter{subfigure}{1}
	\caption{Scenes with pseudo-stationary backgrounds from the ReBaQ dataset.}
	\label{fig:dynReBaQ}
	\end{subfigure}
\caption{Reference background images for different scenes in the Reconstructed Background Quality (ReBaQ) Dataset. Each reference background image corresponds to a captured scene background without foreground objects.}
\label{fig:ReBaQDataset}
\end{figure}

This database consists of sequences of multiple images for eight different scenes. Every image sequence consists of 8 different views such that the background is visible at every pixel in at least one of the views. A reference background image that is free of any foreground objects is also captured for every scene. Figure~\ref{fig:ReBaQDataset} shows the reference images corresponding to each of the eight different scenes in this database.

Each of the image sequences is used as input to twelve different background reconstruction algorithms \cite{SOBS, SpatioTemp, bkgEstimator, BTS, MRF, MultiLayer, PBI, Cormac, Photomontage, CnGrad}. The 144 ($8\times12$) background images generated by these algorithms along with the corresponding reference images for the scene are then used for the subjective evaluation.
Each of the scenes pose a different challenge for the background reconstruction algorithms. 
For example, ``Street'' and ``Wall''  are outdoor sequences with textured backgrounds while the ``Hall'' is an indoor sequence with textured background. 
The ``WetFloor'' sequence challenges the underlying principal of many background reconstruction algorithms with water appearing as a low-contrast foreground object. 
The ``Escalator'' sequence has large motion in the background due to the moving escalator, while ``Park'' has smaller motion in the background due to waving trees. The ``Illumination'' sequence exhibits changing light sources, directions and intensities while the ``Building'' sequence has changing reflections in the background. 
Broadly, the dataset contains two categories based on the scene characteristics: (i)~Static, the scenes for which all the pixels in the background are stationary; and (ii)~Dynamic, the scenes for which there are non-stationary background pixels (e.g., moving escalator, waving trees, varying reflections). 
Four out of the eight scenes in the ReBaQ dataset are categorized as Static and the remaining four are categorized as Dynamic scenes. 
The reference background images corresponding to the static scenes are shown in Figure~\ref{fig:ReBaQDataset}\subref{fig:staticReBaQ}. 
Although there are reflections on the floor in the ``WetFloor'' sequence, it does not exhibit variations at the time of recording and hence is categorized as static background scene.  
The reference background images corresponding to the dynamic background scenes are shown in Figure~\ref{fig:ReBaQDataset}\subref{fig:dynReBaQ}.

\subsubsection{SBMNet based Reconstructed Background Quality (S-ReBaQ) Dataset}
\label{sec:dataset2}

\begin{figure}[tp]
\centering
\begin{singlespace}
	\begin{subfigure}{\textwidth}
		\captionsetup{labelformat=empty, format=default}
		\begin{subfigure}{0.24\textwidth}	
			{\includegraphics[width=4cm,height=2.25cm]{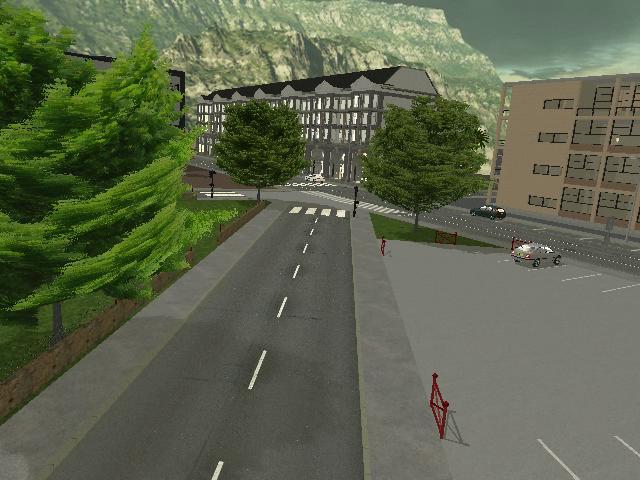}}\hfill%
			\subcaption{511 \\ (Basic)}
		\end{subfigure}
		\begin{subfigure}{0.24\textwidth}
			{\includegraphics[width=4cm,height=2.25cm]{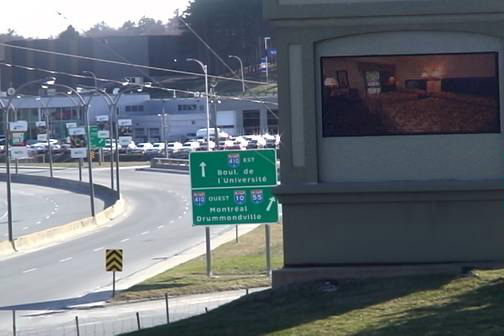}}\hfill%
			\subcaption{\centering Advertisement Board \newline (Background Motion)}
		\end{subfigure}
		\begin{subfigure}{0.24\textwidth}
			{\includegraphics[width=4cm,height=2.25cm]{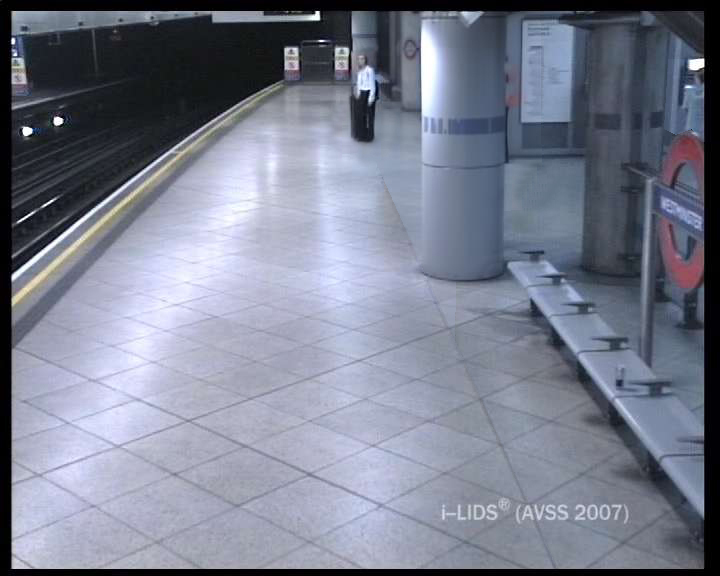}}\hfill%
			\subcaption{\centering AVSS2007 \newline (Intermittent motion)}
		\end{subfigure}
		\begin{subfigure}{0.24\textwidth}
			{\includegraphics[width=4cm,height=2.25cm]{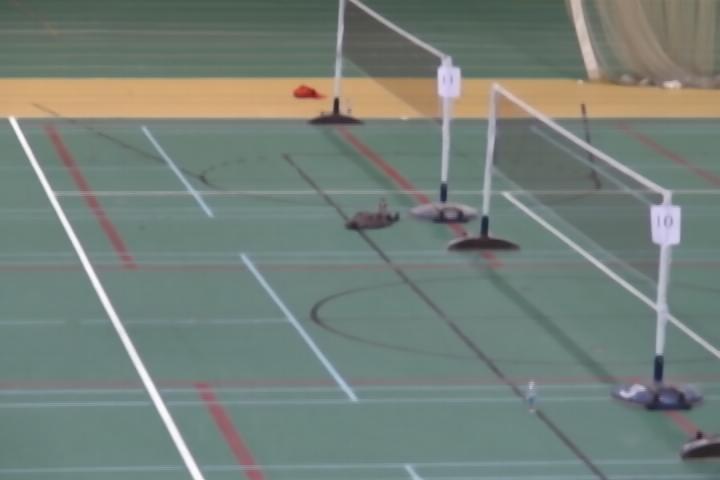}}\hfill%
			\subcaption{\centering Badminton \newline (Jitter)}
		\end{subfigure}
	\end{subfigure} 

	\begin{subfigure}{\textwidth}
		\captionsetup{labelformat=empty, format=default}
		\begin{subfigure}{0.24\textwidth}	
			{\includegraphics[width=4cm,height=2.25cm]{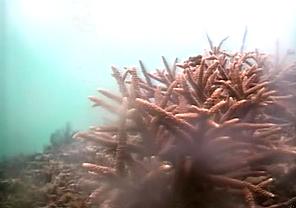}}\hfill%
			\subcaption{Blurred \\ (Basic)}
		\end{subfigure}
		\begin{subfigure}{0.24\textwidth}
			{\includegraphics[width=4cm,height=2.25cm]{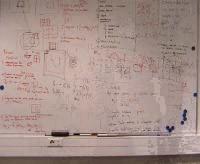}}\hfill%
			\subcaption{Board \\ (Cluttered)}
		\end{subfigure}
		\begin{subfigure}{0.24\textwidth}
			{\includegraphics[width=4cm,height=2.25cm]{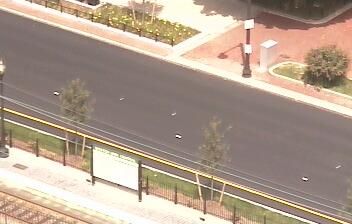}}\hfill%
			\subcaption{Boulevard \\ (Jitter)}
		\end{subfigure}
		\begin{subfigure}{0.24\textwidth}
			{\includegraphics[width=4cm,height=2.25cm]{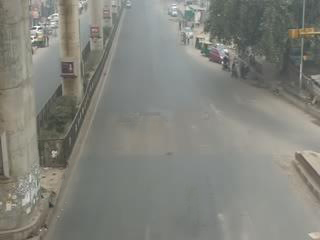}}\hfill%
			\subcaption{\centering Boulevard Jam \newline (Cluttered)}
		\end{subfigure}
	\end{subfigure} 
	
	\begin{subfigure}{\textwidth}
		\captionsetup{labelformat=empty, format=default}
		\begin{subfigure}{0.24\textwidth}	
			{\includegraphics[width=4cm,height=2.25cm]{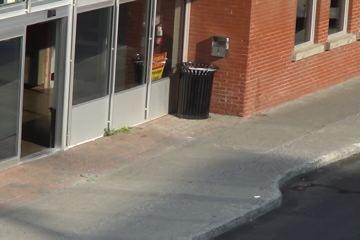}}\hfill%
			\subcaption{\centering Bus Station \newline (Intermittent Motion)}
		\end{subfigure}
		\begin{subfigure}{0.24\textwidth}
			{\includegraphics[width=4cm,height=2.25cm]{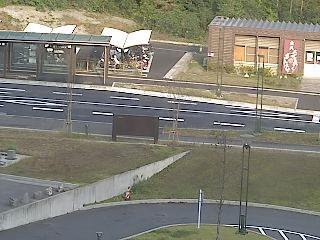}}\hfill%
			\subcaption{\centering Bus Stop in Morning \newline (Long Video)}
		\end{subfigure}
		\begin{subfigure}{0.24\textwidth}
			{\includegraphics[width=4cm,height=2.25cm]{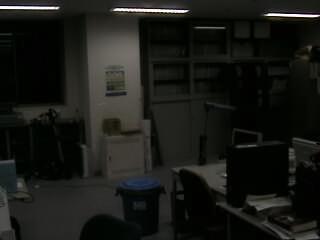}}\hfill%
			\subcaption{\centering Camera Parameter \newline (Illumination Changes)}
		\end{subfigure}
		\begin{subfigure}{0.24\textwidth}
			{\includegraphics[width=4cm,height=2.25cm]{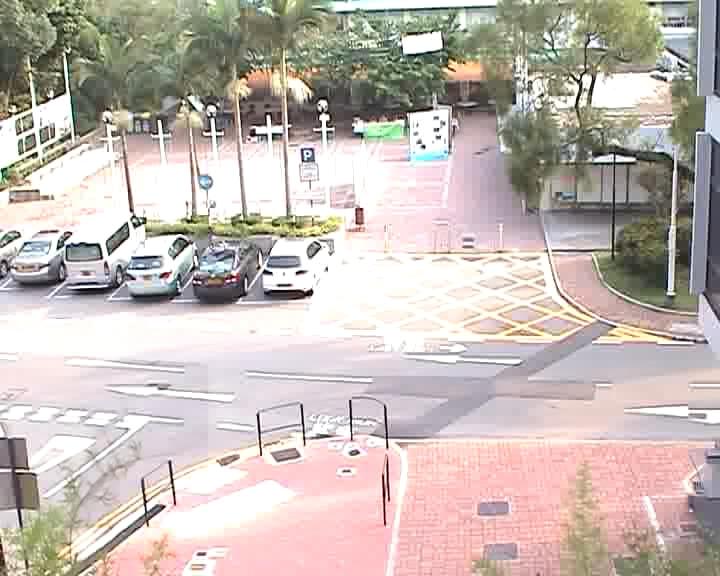}}\hfill%
			\subcaption{\centering CUHK Square  \newline (Short Video)}
		\end{subfigure}
	\end{subfigure} 
	
	\begin{subfigure}{\textwidth}
		\captionsetup{labelformat=empty, format=default}
		\centering 
		\begin{subfigure}{0.24\textwidth}	
			{\includegraphics[width=4cm,height=2.25cm]{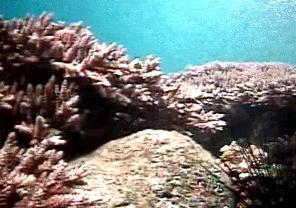}}\hfill%
			\subcaption{\centering Dynamic Background \newline (Short video)}
		\end{subfigure}
	\end{subfigure} 
\caption{Reference background images for different scenes in the SBMNet based Reconstructed Background Quality (S-ReBaQ) Dataset. Each reference background image corresponds to a captured scene background without foreground objects. }
\label{fig:S-ReBaQDataset}
\end{singlespace}
\end{figure}

This dataset is created from the videos in the Scene Background Modeling Net (SBMNet) dataset~\cite{SBMnet} used for the Scene Background Modeling Challenge (SBMC) 2016~\cite{SBMC}. SMBNet consists of image sequences corresponding to a total of 79 scenes. These image sequences are representative of typical indoor and outdoor visual data captured in surveillance, smart environment, and video database scenarios. 
The spatial resolutions of the sequences corresponding to different scenes vary from 240x240 to 800x600. The length of the sequences also varies from 6 to 9,370 images.
The authors of SBMNet categorize these scenes into eight different classes based on the challenges posed~\cite{SBMnet}: 
(a)~Basic category represents a mixture of mild challenges typical of the shadows, Dynamic Background, Camera Jitter and Intermittent Object Motion categories;
(b)~Background motion category includes scenes with strong (parasitic) background motion; for example, in the ``Advertisement Board'' sequence the advertisement board in the scene periodically changes;
(c)~Intermittent Motion category includes sequences with scenarios known for causing ``ghosting'' artifacts in the detected motion;
(d)~Jitter category contains indoor and outdoor sequences captured by unstable cameras;
(e)~Clutter category includes sequences containing a large number of foreground moving objects occluding a large portion of the background;
(f)~Illumination Changes category contains indoor sequences containing strong and mild illumination changes;
(g)~Very Long category contains sequences each with more than 3,500 images;
(h)~Very Short category contains sequences with a limited number of images (less than 20). 
The authors of SBMNet~\cite{SBMnet} provide reference background images for only 13 scenes out of the 79 scenes. 
There is at least one scene corresponding to each category with reference background image available. 
We use only these 13 scenes for which the reference background images are provided. 
Figure~\ref{fig:S-ReBaQDataset} shows the reference background images corresponding to the scenes in this database with the categories from SBMNet~\cite{SBMnet} in brackets.
Background images that were reconstructed by 14 algorithms submitted to SBMC~\cite{Photomontage,MRF,LABGEN_P,SC_SOBS_C4,RPCA,TemporalMedianBI,BE_AAPSA,VotingBI,TMFG,FC_FlowNet,AAPSA,RMR} corresponding to the selected 13 scenes were used in this work for conducting subjective tests. 
As a result, a total of 182 ($13\times14$) reconstructed background images along with their corresponding subjective scores form the S-ReBaQ dataset.

\subsection{Subjective Evaluation}
\label{sec:SubTests}
\begin{figure}[t]
 \centering
  \includegraphics[width=16.5cm]{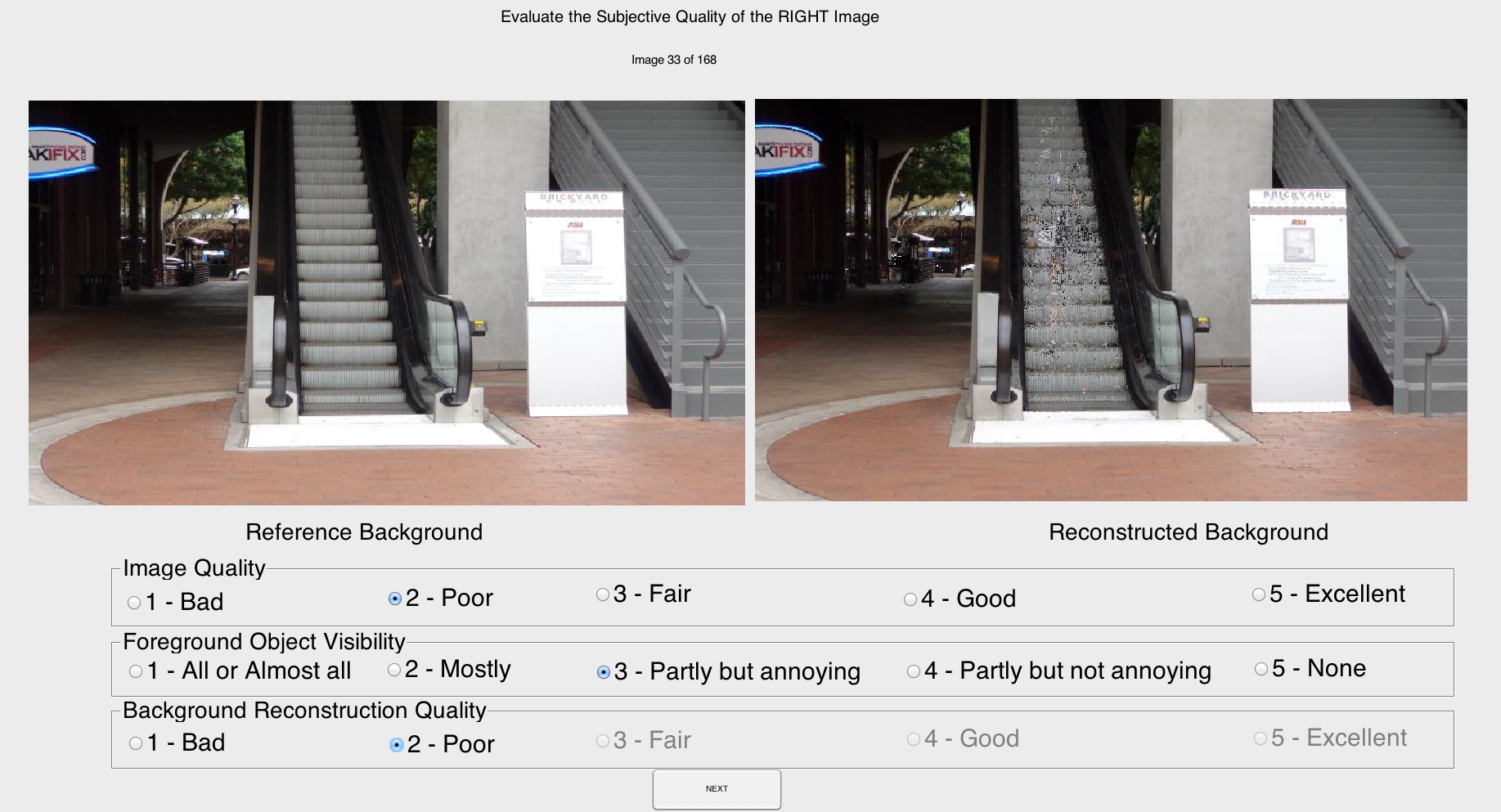}
 \caption{Subjective test Graphical User Interface (GUI).}
\label{fig:GUI}
\end{figure}
\begin{figure}[t]
 \centering
  \includegraphics[width=16.5cm]{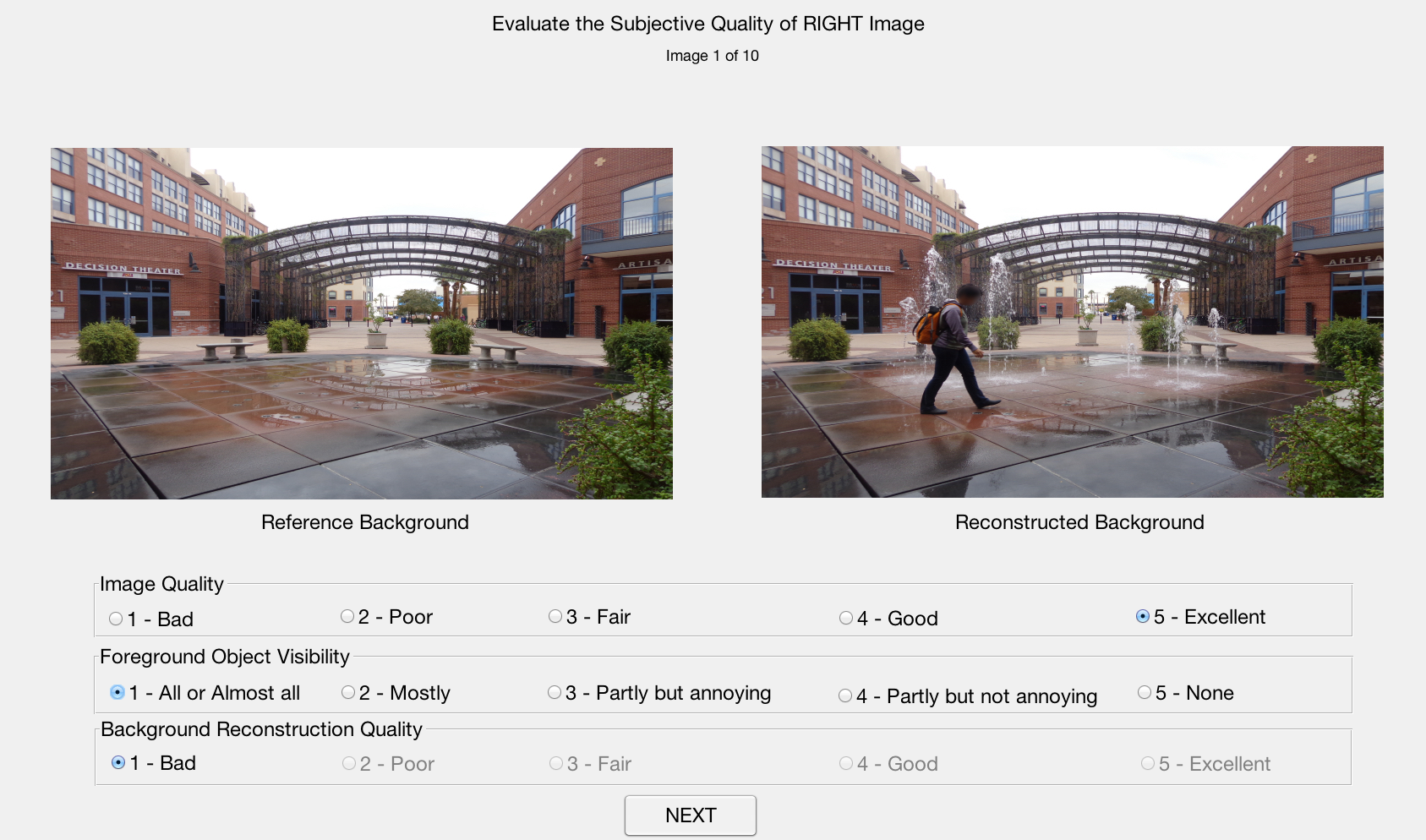}
 \caption{Example of excellent image quality but with all of the foreground visible.}
\label{fig:GUI2}
\end{figure}
The subjective ratings are obtained by asking the human subjects to rate the similarity of the reconstructed background images to the reference background images. 
The subjects had to score the images based on three aspects: 1) overall perceived visual image quality; 2) visibility or presence of foreground objects, and 3) perceived background reconstruction quality.
The subjects had to score the image quality on a 5-point scale, with 1 being assigned to the lowest rating of ‘Bad’ and 5 assigned to the highest rating of ‘Excellent’. 
The second aspect was determining the presence of foreground objects. For our application, we defined the foreground object as any object that is not present in the reference image. The foreground visibility was scored on a 5-point scale marked as: ‘1-All foreground visible’, ‘2-Mostly visible’, ‘3-Partly visible but annoying’, ‘4-Partly visible but not annoying’ and ‘5-None visible’. 
The background reconstruction quality was also measured using a 5-point scale similar to that of the image quality, but the choices were limited based on how the first two aspects of an image were scored. 
For example, as illustrated in Figure~\ref{fig:GUI2}, if the image quality was rated as excellent but the foreground object visibility was rated 1 (all visible), the reconstructed background quality cannot be scored to be very high. The background reconstruction quality scores, referred to as raw scores in the rest of the paper, are used for calculating the Mean Opinion Score (MOS).

We adopted a double-stimulus technique in which the reference and the reconstructed background images were presented side-by-side \cite{ITU} to each subject as shown in Figure~\ref{fig:GUI} and~\ref{fig:GUI2}. Though the same testing strategy and set up was used for the ReBaQ and S-ReBaQ datasets described in Section~\ref{sec:database}, the tests for each dataset were conducted in separate sessions. 

As discussed in \cite{QoMex}, the subjective experiments were carried out on a 23-inch Alienware monitor with a resolution of 1920x1080. Before the experiment, the monitor was reset to its factory settings. The setup was placed in a laboratory under normal office illumination conditions. Subjects were asked to sit at a viewing distance of 2.5 times the monitor height. 

Seventeen subjects participated in the subjective test for the ReBaQ dataset, while sixteen subjects participated in the subjective test for the S-ReBaQ dataset.
The subjects were tested for vision and color blindness using the Snellen chart \cite{Snellen} and Ishihara color vision test \cite{Ishihara}, respectively. A training session was conducted before the actual subjective testing, in which the subjects were shown few images covering different quality levels and distortions of the reconstructed background images and their responses were noted to confirm their understanding of the tests.

\setlength{\belowcaptionskip}{0pt}
\begin{figure}[tp]
  \centering
\begin{subfigure}[t]{\textwidth}
  \hspace*{\fill}%
  \includegraphics[width=4cm]{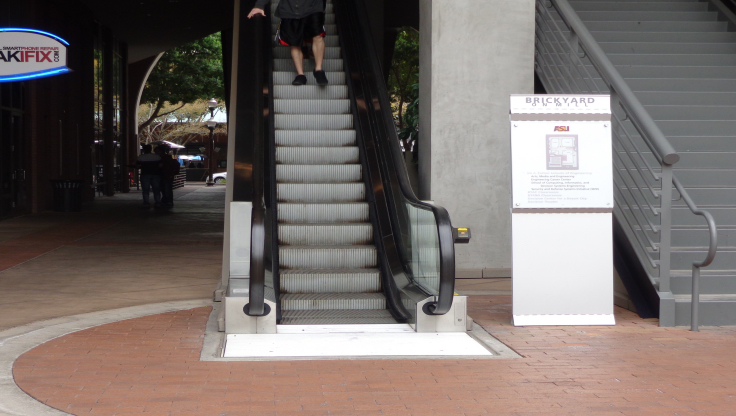}\hfill%
  \includegraphics[width=4cm]{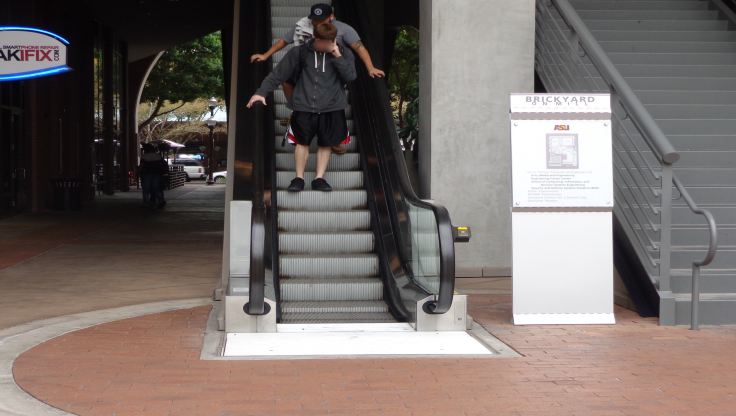}\hfill%
  \includegraphics[width=4cm]{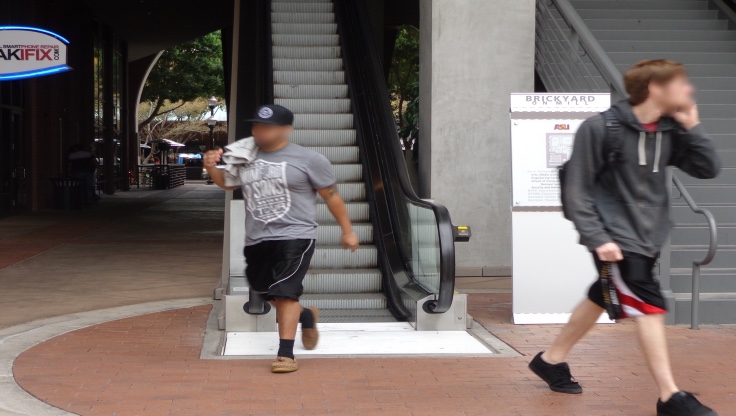}\hfill%
  \includegraphics[width=4cm]{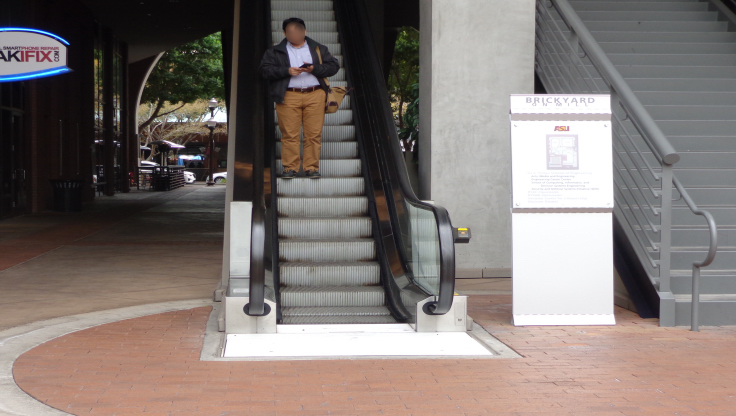}%
  \caption{Four out of eight images from the input sequence "Escalator".}
\label{fig:EscIn}
\end{subfigure}
\vspace*{2mm}
\setlength{\belowcaptionskip}{-5pt}
\begin{subfigure}[t]{\textwidth}
	\begin{subfigure}{0.24\textwidth}
		\captionsetup{labelformat=empty, format=default}
		\includegraphics[width=4cm]{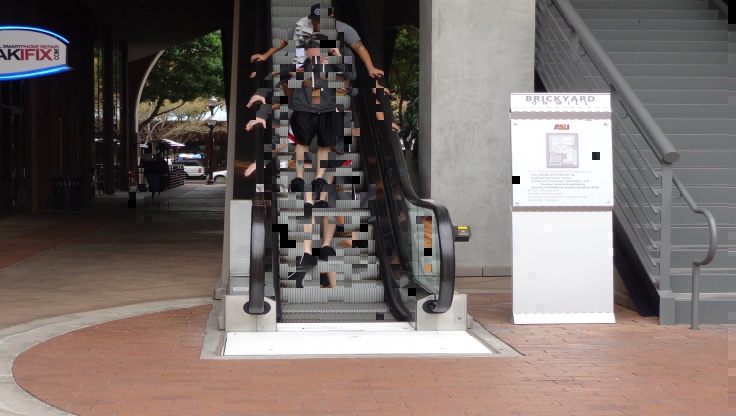}\hfill%
		\subcaption{\cite{BTS}, MOS=1.5882}
	\end{subfigure}
	\begin{subfigure}{0.24\textwidth}
		\captionsetup{labelformat=empty, format=default}
		  \includegraphics[width=4cm]{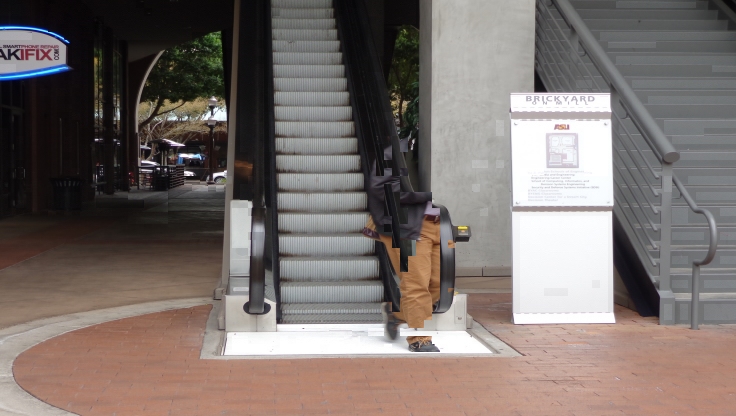}\hfill%
		\subcaption{\cite{MRF}, MOS=2.2353}
	\end{subfigure}
	\begin{subfigure}{0.24\textwidth}
		\captionsetup{labelformat=empty, format=default}
		\includegraphics[width=4cm]{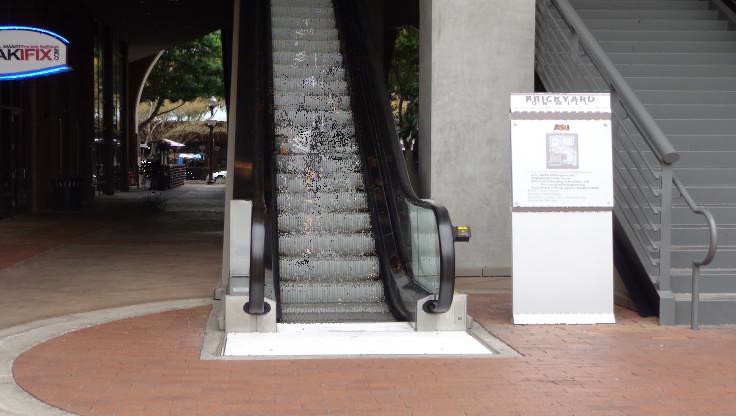}\hfill%
		\subcaption{\cite{MultiLayer}, MOS=2.2941}
	\end{subfigure}
	\begin{subfigure}{0.24\textwidth}
		\captionsetup{labelformat=empty, format=default}
		\includegraphics[width=4cm]{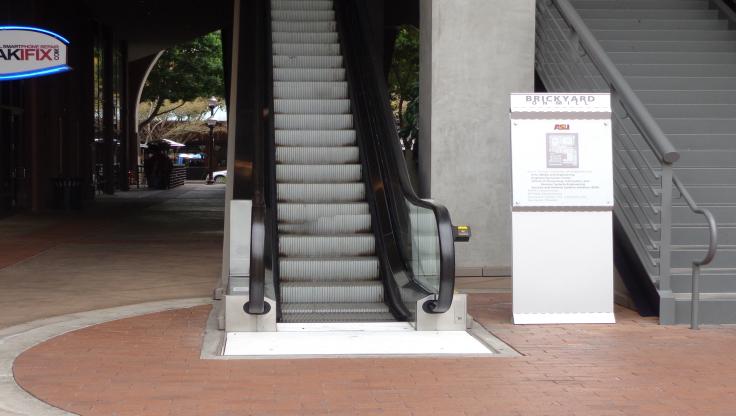}%
		\subcaption{\cite{Photomontage}, MOS=4.1176}
	\end{subfigure}
	 \vspace*{2mm}
  \setcounter{subfigure}{1}
  \caption{Background images reconstructed by different algorithms and corresponding MOS scores.}
  \label{fig:EscOut}
\end{subfigure}

  \caption{Example input sequence and recovered background images with corresponding MOS scores from the ReBaQ dataset.}
  \label{fig:Examples}
\end{figure}
Since the number of participating subjects was less than 20 for each of the datasets, the raw scores obtained by subjective evaluation were screened using the procedure in ITU-R BT 500.13 \cite{ITU}. The kurtosis of the scores is determined as the ratio of the fourth order moment and the square of the second order moment. If the kurtosis lies between 2 and 4, the distribution of the scores can be assumed to be normal. If more than 5\% of the scores given by a particular subject lie outside the range of 2 standard deviations from the mean scores in case of normally distributed scores, that subject is rejected. For the scores that are not normally distributed the range is determined as $\sqrt{20}$ times the standard deviation. In our study two subjects were found to be outliers and the corresponding scores were rejected for the ReBaQ dataset, while no subject was rejected for the S-ReBaQ dataset. MOS scores were calculated as the average of the raw scores retained after outlier removal.
The raw scores and MOS scores with the standard deviations are provided along with the database.

Figure~\ref{fig:Examples} shows an input sequence for a scene in the ReBaQ dataset together with reconstructed background images using different algorithms and corresponding MOS scores. 

\section{Proposed Reconstructed Background Quality Index}
\label{sec:Proposed}
In this section we propose a full-reference quality index that can automatically assess the perceived quality of the reconstructed background images. 
The proposed Reconstructed Background Quality Index (RBQI) uses a probability summation model to combine visual characteristics at multiple scales and quantify the deterioration in the perceived quality of the reconstructed background image due to the presence of any residual foreground objects or unnaturalness that may be introduced by the background reconstruction algorithm. 
The motivation for RBQI comes from the fact that the quality of a reconstructed background image depends on two factors namely: 
\begin{inparaenum}[(i)]
	\item the visibility of the foreground objects, and 
	\item the visible artifacts introduced while reconstructing the background image.
\end{inparaenum}

\begin{figure}[!ht]
\centering
  \includegraphics[width=0.5\linewidth]{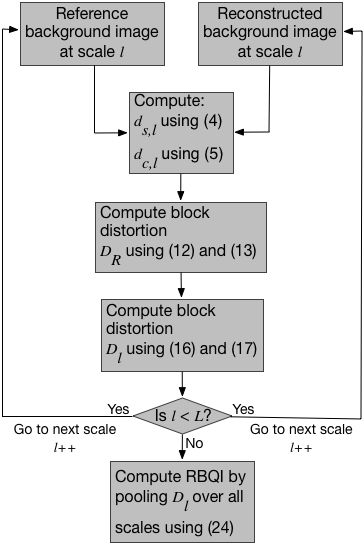}
  \caption{Block diagram describing the computation of the proposed Reconstructed Background Quality Index (RBQI).}
  \label{fig:BlockDiag}
\end{figure}


A block diagram of the proposed quality index (RBQI) is shown in Figure~\ref{fig:BlockDiag}.
An $L$-level multi-scale decomposition of the reference and reconstructed background images is obtained through lowpass filtering using an averaging filter~\cite{MSSIM} and downsampling, where $l=0$ corresponds to finest scale and $l=L-1$ corresponds to the coarsest scale. 
For each level $l=0,...,L-1$, contrast, structure and color differences are computed locally at each pixel to produce a contrast-structure difference map and a color difference map. 
The difference maps are combined in local regions within each scale and later across scales using a `probability summation model' to predict the perceived quality of the reconstructed background image.
More details about the computation of the difference maps and the proposed RBQI based on a probability summation model are provided below. 

\subsection{Structure Difference Map ($d_s$)}
\label{ssec:dsMap}
An image can be decomposed into three different components: luminance, contrast and structure \cite{SSIM}. By comparing these components, similarity between two images can be calculated~\cite{SSIM, MSSIM}. 
A reconstructed background image is formed by mosaicing together parts of different input images, hence, preservation of the local luminance from the reference background image is of low relevance as long as the structure continuity is maintained.
Any sudden variation in the local luminance across the reconstructed background image manifests itself as contrast or structure deviation from the reference image.
Thus, in our application we consider only contrast and structure for comparing the reference and reconstructed background images while leaving out the luminance component. 
These contrast and structure differences between the reference and the reconstructed background images, calculated at each pixel, give us the `contrast-structure difference map' referred to as `structure map' for short in the rest of the paper.

First the structure similarity between the reference and the reconstructed background image, referred to as Structure Index ($SI$), is calculated using~\cite{SSIM}:
\begin{equation}
\label{eq:SI}
	SI(x,y) = \frac{2\sigma_{r_{(x,y)}i_{(x,y)}} + C}{\sigma^2_{r_{(x,y)}} + \sigma^2_{i_{(x,y)}} + C}
\end{equation}
where $r$ is the reference background image, $i$ is the reconstructed background image, $\sigma_r$ and  $\sigma_i$ are the standard deviations of the reference and reconstructed background image, respectively. $\sigma_{r_{(x,y)},i_{(x,y)}}$ is the cross-correlation between the reference and reconstructed background images at location $(x,y)$. $C$ is a small constant to avoid instability and is calculated as $C=(K\cdot l)^2$, $K$ is set to $0.03$ and $l$ is the maximum possible value of the pixel intensity ($255$ in this case)~\cite{SSIM}.
A higher $SI$ value indicates higher similarity between the pixels in the reference and reconstructed background images. 

The background scenes often contain pseudo-stationary objects such as waving trees, escalator, local and global illumination changes. Even though these pseudo-stationary pixels belong to the background, because of the presence of motion, they are likely to be classified as foreground pixels. 
For this reason the pseudo-stationary backgrounds pose an additional challenge for the quality assessment algorithms. 
Just comparing co-located pixel neighborhoods in the two considered images is not sufficient in the presence of such dynamic backgrounds, our algorithm uses a search window of size $nhood \times nhood$ centered at the current pixel $(x,y)$ in the reconstructed image, where $nhood$ is an odd value. 
The $SI$ is calculated between the pixel at location $(x,y)$ in the reference image and $(nhood)^2$ pixels within the $nhood \times nhood$ search window centered at pixel $(x,y)$ in the reconstructed image. 
The resulting $\boldsymbol{SI}$ matrix is of size $nhood \times nhood$ .
The modified Equation~(\ref{eq:SI}) to calculate $SI$ for every pixel location in the $nhood \times nhood$ window centered at $(x,y)$ is given as:
\begin{equation}
\label{eq:SImn}
	\boldsymbol{SI}_{(x,y)}(m,n) = \frac{2\sigma_{r_{(x,y)}i_{(m,n)}} + C}{\sigma^2_{r_{(x,y)}} + \sigma^2_{i_{(m,n)}} + C} 
\end{equation}
where
\begin{align}
	 m = x - (nhood-1)/2 : x + (nhood-1)/2 \nonumber \\
           n = y - (nhood-1)/2 : y + (nhood-1)/2 \nonumber
\end{align}
The maximum value of the $\boldsymbol{SI}$ matrix is taken to be the final $SI$ value for the pixel at location $(x,y)$ as given below:
\begin{equation}
	SI(x,y) = \max_{(m,n)}(\boldsymbol{SI}_{(x,y)}(m,n))
\end{equation}
The $SI$ map takes on values between [-1,1]. 
\begin{figure*}[t]
\centering
\begin{subfigure}[t]{.49\linewidth}
  \centering
 \frame{\includegraphics[width=1.0\linewidth]{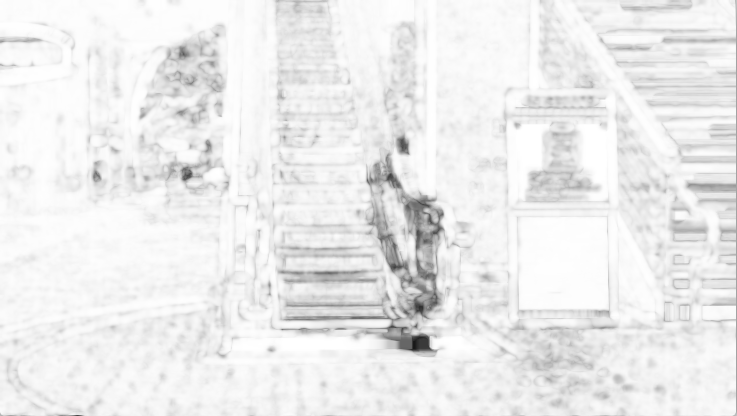}}
	\caption{scale $l$=0}
\end{subfigure}
\hfill
\begin{subfigure}[t]{.29\linewidth}
  \centering
  \frame{\includegraphics[width=1.0\linewidth]{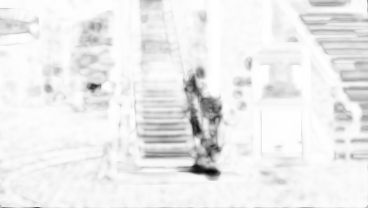}}
	\caption{scale $l$=1}
\end{subfigure}
\hfill
\begin{subfigure}[t]{.19\linewidth}
  \centering
\frame{\includegraphics[width=1.0\linewidth]{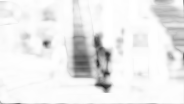}}
	\caption{scale $l$=2}
\end{subfigure} \\

\caption{Quality map with $nhood=17$ for the background image shown in Figure~\ref{fig:Examples}\subref{fig:EscOut} and reconstructed using the method in \cite{MRF}. The darker regions indicate larger structure differences between the reference and the reconstructed background images.}
\label{fig:EscMap}
\end{figure*}
In the proposed method, the $SI$ map is computed at $L$ different scales denoted as $SI_l(x,y), l=0,...,L-1$.
The quality maps generated at three different scales for the background image shown in Figure~\ref{fig:Examples}\subref{fig:EscOut} and reconstructed using method in \cite{MRF} are shown in Figure~\ref{fig:EscMap}.
The darker regions in these images indicate larger structure differences between the reference and the reconstructed background images while the lighter regions indicate higher similarities.

The structure difference map is calculated using the $SI$ map at each scale $l$ as follows:
\begin{equation}
\label{eq:dsMap}
	d_{s,l}(x,y) = \frac{1- SI_l(x,y)}{2}
\end{equation}
$d_{s,l}$ takes on values between [0,1] where the value of $0$ corresponds to no difference while $1$ corresponds to largest difference. 

\subsection{Color Distance ($d_c$)}
\label{ssec:dcMap}
The $d_{s,l}$ map is vulnerable to failures while detecting differences in areas of background images with no textures or no structural information and/or with objects of same luminance but different color. 
Hence we incorporate the color information at every scale while calculating the RBQI. 
The reference and the reconstructed images are converted to the $Lab$ color space and filtered using a lowpass Gaussian filter. The color difference between the filtered reference and reconstructed background images at each scale $l$ is then calculated as the Euclidian distance between the values of co-located pixels as follows:
\begin{equation}
\label{eq:dcMap}
	d_{c,l}(x,y) = \sqrt{(L_{r}(x,y)-L_{i}(x,y))^2 + (a_{r}(x,y)-a_{i}(x,y))^2 + (b_{r}(x,y)-b_{i}(x,y))^2}
\end{equation}
In~\eqref{eq:dcMap}, for the $Lab$ color space components the scale index $l$ was dropped from the notation for convenience.

\subsection{Computation of the Reconstructed Background Quality Index (RBQI) based on Probability Summation}
\label{ssec:RBQI}
As indicated previously, the reference and reconstructed background images are decomposed each into a multi-scale pyramid with $L$ levels. 
Structure difference maps $d_{s,l}$ and color difference maps $d_{c,l}$ are computed at every level $l=0,...,L-1$ as described in Equations~\eqref{eq:dsMap} and~\eqref{eq:dcMap}, respectively.
These difference maps are pooled together within the scale and later across all scales using a probability summation model~\cite{psychometric} to give the final RBQI.

The probability summation model as described in~\cite{psychometric} considers an ensemble of independent difference detectors at every pixel location in the image. 
These detectors predict the probability of perceiving the difference between the reference and the reconstructed background images at the corresponding pixel location based on its neighborhood characteristics in the reference image. 
Using this model, the probability of the structure difference detector signaling presence of a structure difference at pixel location $(x,y)$ at level $l$ can be modeled as an exponential of the form: 
\begin{equation}
\label{eq:PDsl}
	P_{D,s,l}(x,y) = 1 - \exp\Bigg(- \Bigg|{\frac{d_{s,l}(x,y)}{\alpha_{s,l}(x,y)}}\Bigg|^{\beta_s}\Bigg)
\end {equation}
where $\beta_s$ is a parameter chosen to increase the correspondence of RBQI with experimentally determined MOS scores on a training dataset and $\alpha_{s,l}(x,y)$ is a parameter whose value depends upon the texture characteristics of the neighborhood centered at $(x,y)$ in the reference image. 
The value of $\alpha_{s,l}(x,y)$ is chosen to take into account that differences in structure are less perceptible in textured areas as compared to non-textured areas.

In order to determine the value of $\alpha$, every pixel in the reference image is classified as textured or non-textured using the technique in~\cite{EdgeClassification}. 
This method first calculates the local variance at each pixel using a 3x3 window centered around it. 
Based on the computed variances a pixel is classified as edge, texture or uniform. 
By considering the number of edge, texture and uniform pixels in the 8x8 neighborhood of the pixel, it is further classified into one of the six types: uniform, uniform/texture, texture, edge/texture, medium edge and strong edge. 
For our application we label the pixels classified as `texture' and `edge/texture' as 'textured' pixels and we label the rest as `non-textured' pixels. 
 
Let, $f_{tex}(x,y)=1$ be the flag indicating that a pixel is textured. Thus values of $\alpha_{s,l}(x,y)$ can be expressed as:
\begin{equation}
	\alpha_{s,l}(x,y) = 
	\begin{cases}
		1.0, & \text{if } f_{tex}(x,y) = 0 \\
		a, \text{where } a \gg 1.0, & \text{if } f_{tex}(x,y) = 1
	\end{cases}
\end{equation}
In our implementation we chose the value of $a=1000.0$ resulting in a value of $P_{D,s,l}$ close to zero when a pixel is classified as textured.

Similarly, the probability of the color difference detector signaling the presence of a color difference at pixel location $(x,y)$ at level $l$ can be modeled as:
\begin{equation}
\label{eq:PDcl}
	P_{D,c,l}(x,y) = 1-\exp\Bigg(- \Bigg|{\frac{d_{c,l}(x,y)}{\alpha_{c,l}(x,y)}}\Bigg|^{\beta_c}\Bigg)
\end {equation}
where $\beta_c$ is found in a similar way to $\beta_s$ and $\alpha_{c,l}(x,y)$ corresponds to the Adaptive Just Noticeable Distortion (AJNCD) calculated at every pixel $(x,y)$ in the $Lab$ color space as given in~\cite{AJNCD}:
\begin{equation}
\label{eq:alpha_c}
	\alpha_{c,l}(x,y) = JNCD_{Lab} \cdot s_L(E(L_l(x,y)),\Delta{L_l(x,y)}) \cdot s_C(a_l(x,y),b_l(x,y)) 
\end{equation}
where $JNCD_{Lab}$ is set to 2.3~\cite{JNCD}, $E(L_l)$ is the mean background luminance of the pixel at $(x,y)$ and $\Delta L$ is the maximum luminance gradient across pixel $(x,y)$.
In Equation~\eqref{eq:alpha_c}, $s_C$ is the scaling factor used for adjusting the dimension of ellipsoid along the chroma axis as is given by~\cite{AJNCD}:
\begin{equation}
	s_C(a_l(x,y),b_l(x,y)) = 1 + 0.045 \cdot (a_l^2(x,y) + b_l^2(x,y))^{1/2} 
\end{equation}
where $a_l(x,y)$ and $b_l(x,y)$ correspond to the $a$ and $b$ color values of pixel located at $(x,y)$ in the Lab color space, respectively.
$s_L$ is the scaling factor that simulates the local luminance texture masking as is given by:
\begin{equation}
	s_L(E(L_l),\Delta{L_l}) = \rho(E(L_l))\Delta{L_l} + 1.0
\end{equation}
where $\rho(E(L_l))$ is the weighting factor as described in~\cite{AJNCD}.
Thus, $\alpha_{c,l}$ varies at every pixel location based on the distance between the chroma values and texture masking properties of its neighborhood.

A pixel $(x,y)$ at the $l$-th level is said to have no distortion if and only if neither the structure difference detector nor the color difference detector at location $(x,y)$ signal the presence of any differences. 
Thus, the probability of detecting no difference between reference and reconstructed background images at pixel $(x,y)$ and level $l$ can be written as:
\begin{equation}
\label{eq:Pndl}
	P_{ND,l}(x,y) = (1-P_{D,s,l}) \cdot (1- P_{D,c,l}) 
\end{equation}
Substituting Equation~\eqref{eq:PDsl} and Equation~\eqref{eq:PDcl} for $P_{D,s,l}$ and $P_{D,c,l}$, respectively, in the above equation, we get:
\begin{equation}
	P_{ND,l}(x,y) = \exp\Bigg(-\Bigg|{\frac{d_{s,l}(x,y)}{\alpha_{s,l}(x,y)}}\Bigg|^{\beta_s}\Bigg) \cdot \exp\Bigg(-\Bigg|{\frac{d_{c,l}(x,y)}{\alpha_{c,l}(x,y)}}\Bigg|^{\beta_c}\Bigg) 
\end{equation}

A less localized probability of difference detection can be computed by adopting the probability summation hypothesis which pools over the localized detection probabilities over a region $R$~\cite{psychometric}. The probability summation hypothesis is based on the following two assumptions: 1) no difference is detected if none of the detectors in the region $R$ sense the presence of distortion, and 2) the probabilities of detection at all locations in the region $R$ are independent. 
Then the probability of no difference detection over the region $R$ is given by:
\begin{equation}
\label{eq:PND_R}
	P_{ND,l}(R) = \prod_{(x,y) \in R}  P_{ND,l}(x,y) 
\end{equation}
Substituting Equation~\eqref{eq:Pndl} in the above equation gives:
\begin{equation}
	P_{ND,l}(R) = \exp\big(- D_{s,l}(R)^{\beta_s} \big)  \cdot \exp\big(- D_{c,l}(R)^{\beta_c} \big) 
\end{equation}
where
\begin{align}
\label{eq:Dlr}
	D_{s,l}(R) = \Bigg(\sum_{(x,y) \in R} \Bigg|{\frac{d_{s,l}(x,y)}{\alpha_{s,l}(x,y)}}\Bigg|^{\beta_s} \Bigg)^{\frac{1}{\beta_s}}  \\ 
	D_{c,l}(R) = \Bigg(\sum_{(x,y) \in R} \Bigg|{\frac{d_{c,l}(x,y)}{\alpha_{c,l}(x,y)}}\Bigg|^{\beta_c} \Bigg)^{\frac{1}{\beta_c}} 
\end{align}
In the human visual system, the highest visual acuity is limited to the size of foveal region, which covers approximately $2^\circ$ of visual angle. In our work, we consider the image regions $R$ as foveal regions approximated by $8\times8$ non-overlapping image blocks. 

The probability of no distortion detection over the $l$-th level is obtained by pooling the no detection probabilities over all the regions $R$ and is given by:
\begin{equation}
	P_{ND}(l) = \prod_{R \in l} P_{ND,l}(R)
\end{equation}
or
\begin{equation}
	P_{ND}(l) = \exp\big(- D_{s}(l)^{\beta_s} \big)  \cdot \exp\big(- D_{c}(l)^{\beta_c} \big) 
\end{equation}
where
\begin{align}
\label{eq:Dl}
	D_{s}(l) = \bigg(\sum_{R \in l} D_{s,l}(R)^{\beta_s} \bigg)^{\frac{1}{\beta_s}}  \\ 
	D_{c}(l) = \bigg(\sum_{R \in l} D_{c,l}(R)^{\beta_c} \bigg)^{\frac{1}{\beta_c}} 
\end{align}
Thus the final probability of detecting no distortion in a reconstructed background image $i$ is obtained by pooling the no detection probabilities $P_ND(l)$ over all scales $l$, $l=0,...,L-1$, as follows:
\begin{equation}
	P_{ND}(i)= \prod_{l=0}^{l=L-1} P_{ND}(l) 
\end{equation}
or
\begin{equation}
\label{eq:Pnd}
	P_{ND}(i) = \exp\big(- D_{s}^{\beta_s} \big)  \cdot \exp\big(- D_{c}^{\beta_c} \big) 
\end{equation}
where
\begin{align}
\label{eq:D}
	D_{s} = \bigg(\sum_{l=0}^{l=L-1} D_{s}(l)^{\beta_s} \bigg)^{\frac{1}{\beta_s}} \\ 
	D_{c} = \bigg(\sum_{l=0}^{l=L-1} D_{c}(l)^{\beta_c} \bigg)^{\frac{1}{\beta_c}} 
\end{align}
From Equation~\eqref{eq:D}, it can be seen that $D_s$ and $D_c$ take the form of a Minkowski metric with exponent $\beta_s$ and $\beta_c$, respectively. 
 
By substituting the values $D_s$, $D_c$, $D_s(l)$, $D_c(l)$, $D_{s,l}(R)$ and $D_{c,l}(R)$ in Equation~\eqref{eq:Pnd} and simplifying, we get:
\begin{equation}
	P_{ND}(i) = \exp( - D)
\end{equation}
where 
\begin{equation}
\label{eq:finalD}
	D = \Big(\sum_{l=0}^{l=L-1} \sum_{R \in l} \sum_{(x,y) \in R} \big[ D_{s,l}(x,y) + D_{c,l}(x,y) \big] \Big)
\end{equation}
Thus the probability of detecting a difference between the reference image and a reconstructed background image $i$ is given as:
\begin{align}
\label{eq:Pd}
	P_{D}(i) = 1 - P_{ND}(i) =  1 - \exp( - D)
\end{align}
As it can be seen from Equation~\eqref{eq:Pd}, a lower value of $D$ results in a lower probability of difference detection $P_{D}(i)$ while a higher value results in a higher probability of difference detection.
Therefore, $D$ can be used to assess the perceived quality in the reconstructed background image, with a lower value of $D$ corresponding to a higher perceived quality.

The final Reconstructed Background Quality Index (RBQI) for a reconstructed background image is calculated using the logarithm of $D$ as follows:
\begin{equation}
	RBQI =  \log_{10}(1 + D)
\end{equation}
As $D$ increases the value of RBQI increases implying more perceived distortion and thus lower quality of the reconstructed background image.
The logarithmic mapping models the saturation effect, i.e., beyond a certain point the maximum annoyance level is reached and more distortion does not affect the quality.

\section{Results}
\label{sec:results}
In this section we analyze the performance of RBQI in terms of its ability to predict the subjective ratings for the perceived quality of reconstructed background images. We evaluate the performance of the proposed quality index in terms of its prediction accuracy, prediction monotonicity and prediction consistency and provide comparisons with the existing statistical and IQA techniques. 
In our implementation, we set $nhood=17$, $L=3$ and $\beta_s=\beta_c=3.5$.
We also evaluate the performance of RBQI for different scales and neighborhood search windows. 
We conduct a series of hypothesis tests based on the prediction residuals (errors in predictions) after nonlinear regression. These tests help in making statistically meaningful conclusions on the index's performance. 

We use the two databases ReBaQ and S-ReBaQ described in Section~\ref{sec:database} to quantify and compare the performance of RBQI.
For performance evaluation, we employ three most commonly used metrics: (i) Spearman rank-order correlation coefficient; (ii) Pearson correlation coefficient; and (iii) root mean squared error (RMSE). 
A 4-parameter regression function \cite{VQEG2000} is applied to IQA metrics to provide a non-linear mapping between the objective scores and the subjective mean opinion scores (MOS):
\begin{equation}
	MOS_{p_{i}} = \frac{\gamma_{1} - \gamma_{2}}{1 + e^{-\left( \frac{M_{i}-\gamma_{3}}{|\gamma_{4}|} \right)}} + \gamma_{2}	
\label{eqn:pMOS}
\end{equation}
where $M_i$ denotes the predicted quality for the $i$th image and $MOS_{p_{i}}$ denotes the quality score after fitting, and $\gamma_{n}, n = 1,2,...,4$, are the regression model parameters.

\begin{figure*}[t]
\centering
\begin{subfigure}{.5\linewidth}
  \centering
  \includegraphics[width=0.98\linewidth]{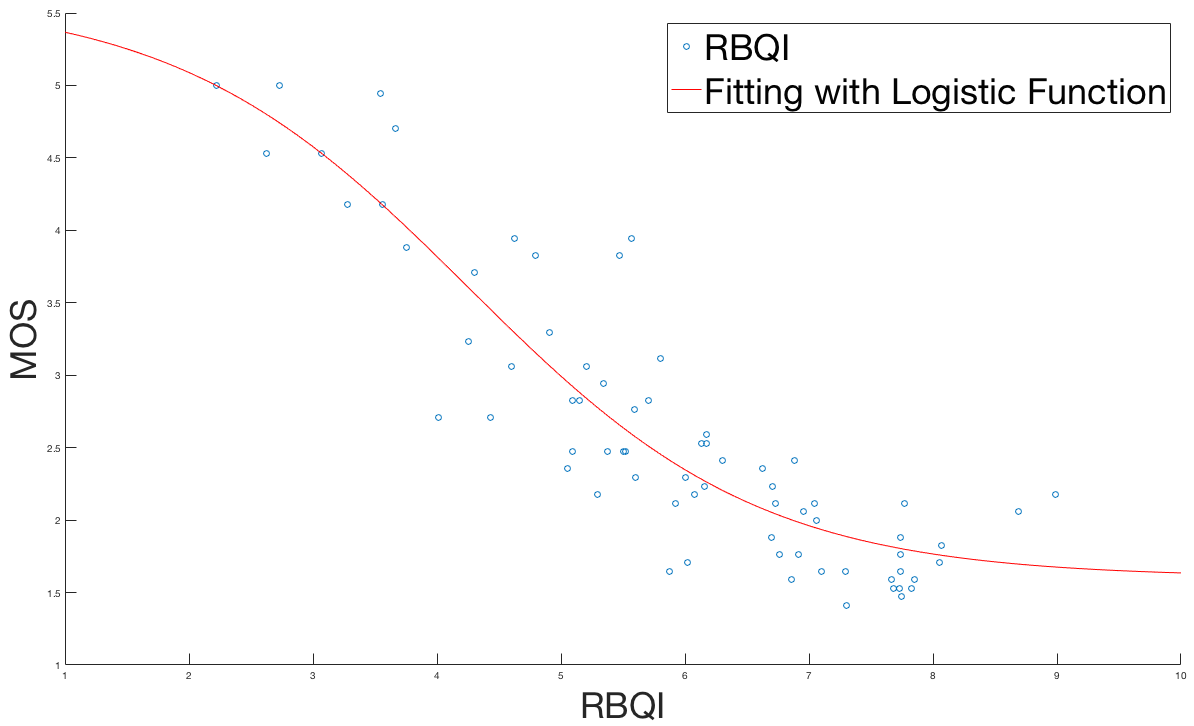}
	\caption{ReBaQ-Static}
  \label{fig:nhood16Fit_static}
\end{subfigure}%
\begin{subfigure}{.5\linewidth}
  \centering
  \includegraphics[width=0.98\linewidth]{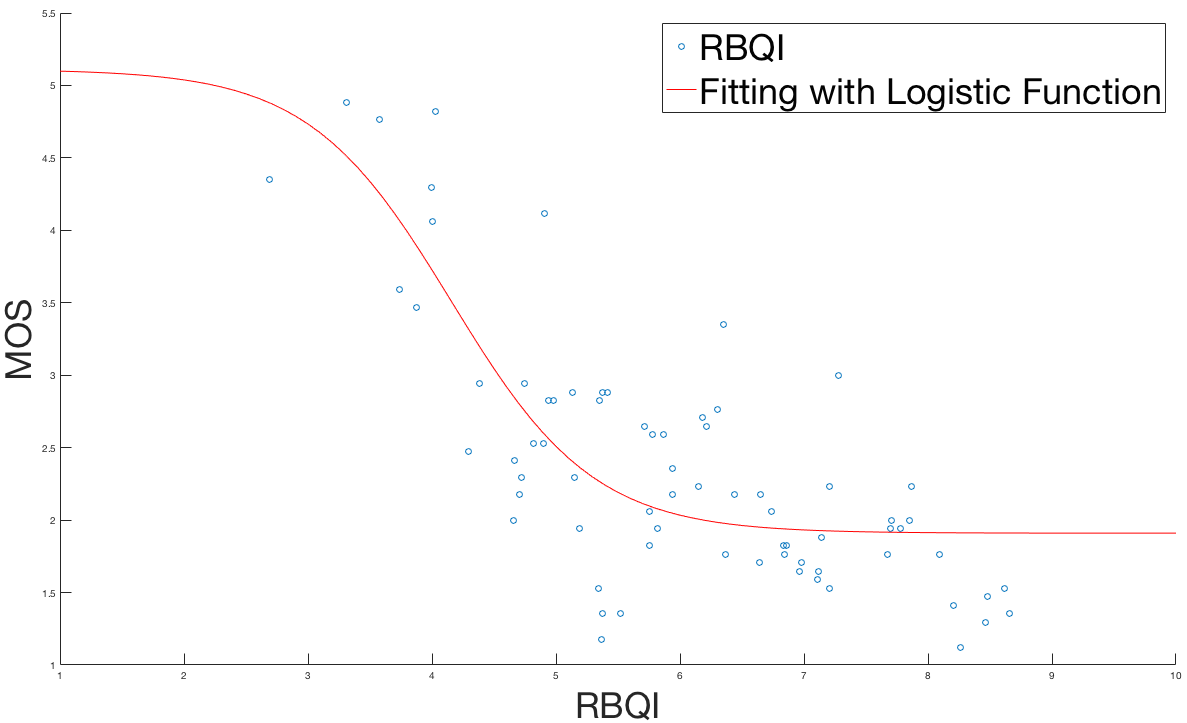}
	\caption{ReBaQ-Dynamic}
  \label{fig:nhood16Fit_dyn}
\end{subfigure} \\ 
\begin{subfigure}{.5\linewidth}
  \centering
  \includegraphics[width=0.98\linewidth]{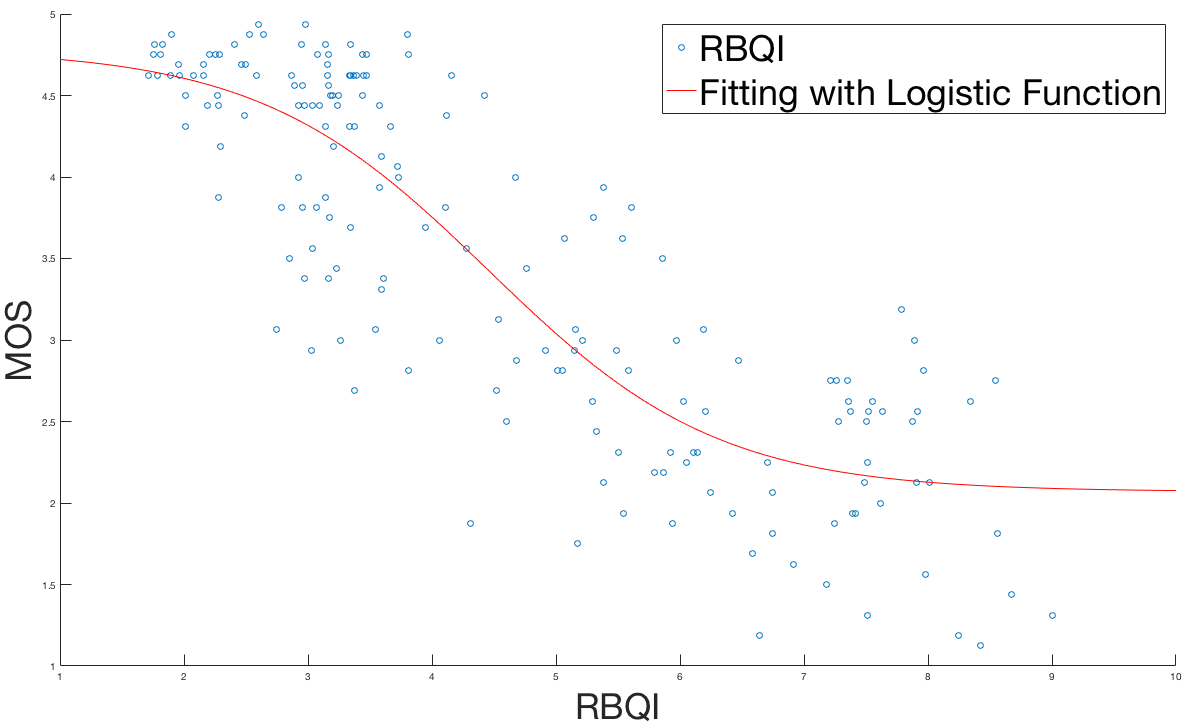}
	\caption{S-ReBaQ}
  \label{fig:nhood16Fit_SBM}
\end{subfigure} \\ 

\caption{Scatter plots of the MOS vs. Metric Scores on different datasets.}
\label{fig:ScatterProposed}
\end{figure*}

Figure~\ref{fig:ScatterProposed} shows the scatter plots of MOS versus the prediction scores using the proposed technique along with the corresponding fitting curve calculated using \eqref{eqn:pMOS}.

\subsection{Performance Comparison}
\label{subsec:PerfComp}

\begin{table}[t]
\centering
\caption{Comparison of RBQI vs. Statistical measures and IQA techniques on the ReBaQ dataset.}
\label{table:ReBaQEval}
\centering
\scriptsize
\renewcommand{\arraystretch}{1.2}
\begin{adjustbox}{center}
\begin{tabular}{|ccc|c|c|c|c||c|c|c|c|c|}
\hline
\multicolumn{ 2}{|c||}{} &  \multicolumn{5}{c||}{\bf{ReBaQ-Static}} 
	& \multicolumn{5}{c|}{\bf{ReBaQ-Dynamic}} \\
\hline
 \multicolumn{ 2}{|c||}{} & {\bf PCC} & {\bf SROCC} & {\bf RMSE} &  {\bf P$_{\text{PCC}}$} & {\bf P$_{\text{SROCC}}$}  
			             & {\bf PCC} & {\bf SROCC} & {\bf RMSE} &  {\bf P$_{\text{PCC}}$} & {\bf P$_{\text{SROCC}}$} \\
\hline \hline
\parbox[t]{6mm}{\multirow{3}{*}{\rotatebox[origin=c]{90}{\tabular{@{}l}Statistical \\ Measures\endtabular}}} 
& \multicolumn{ 1}{|l||} {\bf AGE} &  0.7776  & 0.6348  & 0.6050 & 0.000000 & 0.000000
						 &  0.4999 & 0.2303 & 0.7644 & 0.005000 & 0.051600 \\

&  \multicolumn{ 1}{|l||} {\bf EPs} & 0.3976  & 0.5093  & 0.8829 & 0.000000 & 0.000000
						& 0.1208	& 0.2771 & 0.8761 & 0.007600 & 0.018500 \\

& \multicolumn{ 1}{|l||} {\bf pEPs} &  0.8058  & 0.6170  & 0.5698 & 0.000000 & 0.000000
						  & 0.4734 & 0.2771 & 0.8825 & 0.007600 & 0.018500 \\

& \multicolumn{ 1}{|l||} {\bf CEPs} &  0.5719 & 0.6939  & 0.7893 & 0.000000 & 0.000000
						   & 0.5951 & 0.7549 & 0.7092 & 0.000000 & 0.000000\\

& \multicolumn{ 1}{|l||} {\bf pCEPs} & 0.6281 & 0.7843 & 0.9622 & 0.000000 & 0.000000
						     & 0.6418 & \bf{0.7940} & 0.8826 & 0.000000 & 0.000000 \\

\hline

\parbox[t]{4mm}{\multirow{3}{*}{\rotatebox[origin=c]{90}{\tabular{@{}l}Image Quality Assessment Metrics \endtabular}}} 
& \multicolumn{ 1}{|l||} {\bf PSNR}  & 0.8324 & 0.7040 & 0.5331 & 0.000000 & 0.000000 
						     & 0.5133  & 0.4179  & 0.7575 & 0.000004 & 0.000263\\ 

& \multicolumn{ 1}{|l||} {{\bf SSIM}\scriptsize{\cite{SSIM}}} & 0.5914 & 0.5168 & 0.7759 & 0.000000 & 0.000177
											& 0.0135 & 0.0264 & 0.8826 & 0.910238 & 0.822439\\

& \multicolumn{ 1}{|l||} {\scriptsize{\bf MS-SSIM}\scriptsize{\cite{MSSIM}}} & 0.7230 & 0.7085 & 0.6648 & 0.000000 & 0.000000
											     & 0.5087  & 0.4466  & 0.7598 & 0.000005 & 0.000085 \\

& \multicolumn{ 1}{|l||} {{\bf VSNR}\scriptsize{\cite{VSNR}}} & 0.5216 & 0.3986 & 0.8209 & 0.000003 & 0.000531
										 	  & 0.5090  & 0.1538  & 0.7597 & 0.000005 & 0.198310  \\

& \multicolumn{ 1}{|l||}  {{\bf VIF}\scriptsize{\cite{VIF}}} &  0.3625 & 0.0843 & 0.8968 & 0.001754 & 0.484429
										 & 0.3103 & 0.3328 & 0.8390 & 0.199921 & 0.236522\\

& \multicolumn{ 1}{|l||} {{\bf VIFP}\scriptsize{\cite{VIF}}} & 0.5122 & 0.3684 & 0.8265 & 0.000004 & 0.001470 
										    & 0.4864  & 0.1004  & 0.7711 & 0.000015 & 0.403684\\

&  \multicolumn{ 1}{|l||} {{\bf UQI}\scriptsize{\cite{UQI}}} & 0.6197 & 0.7581 & 0.9622 & 0.000000 & 0.000000
										    & 0.6262  & \bf{0.7450}  & 0.8826 & 0.000000 & 0.000000 \\

&  \multicolumn{ 1}{|l||} {{\bf IFC}\scriptsize{\cite{IFC}}} & 0.5003 & 0.3771 & 0.8331 & 0.000008 & 0.001105 
										   & 0.4306  & 0.1024  & 0.7966 &0.000160 & 0.394409 \\

& \multicolumn{ 1}{|l||} {{\bf NQM}\scriptsize{\cite{NQM}}} & 0.8251 & \bf{0.8602} & 0.5437 & 0.000000 & 0.000000
										      & 0.6898  & 0.6600  & 0.6390 & 0.000000 & 0.000000  \\

& \multicolumn{ 1}{|l||} {{\bf WSNR}\scriptsize{\cite{WSNR}}} & 0.8013 & 0.7389 & 0.5756 &  0.000000 & 0.000000 
										 	    & 0.6409  & 0.5760  & 0.6775 &  0.000000 & 0.000000 \\

& \multicolumn{ 1}{|l||} {{\bf FSIM}\scriptsize{\cite{FSIM}}} & 0.7209 & 0.6970 & 0.6668 &  0.000000 & 0.000000 
										        & 0.5131  & 0.3283  & 0.7575 & 0.000004	& 0.004922 \\

& \multicolumn{ 1}{|l||} {{\bf FSIMc}\scriptsize{\cite{FSIM}}} & 0.7274 & 0.7033 & 0.6603 &  0.000000 & 0.000000
											& 0.5144  & 0.3310  & 0.7568 & 0.000004 & 0.004559 \\

&\multicolumn{ 1}{|l||}  {{\bf SRSIM}\scriptsize{\cite{SRSIM}}} & 0.7906 & 0.7862 & 0.5892 &  0.000000 & 0.000000 
											     & 0.5512  & 0.5376  & 0.7364 & 0.000001 & 0.000001 \\

& \multicolumn{ 1}{|l||} {{\bf SalSSIM}\scriptsize{\cite{SalSSIM}}} & 0.5983 & 0.5217 & 0.7710 & 0.000000 & 0.000003 
												 & 0.4866  & 0.3200  & 0.7710 & 0.000015 & 0.006198 \\

&  \multicolumn{ 1}{|l||} {{\bf CQM}\scriptsize{\cite{CQM}}} & 0.6401 & 0.5755 & 0.7393 & 0.000000 & 0.000000 
											& 0.7050  & 0.7610  & 0.6259 & 0.000000 & 0.000000 \\

& \multicolumn{ 1}{|l||} {\bf RBQI} & \bf{ 0.9006} & 0.8592 & \bf{0.4183} & 0.000000 & 0.000000 & \bf{ 0.7908} & 0.6773 & \bf{0.5402}  & 0.000000 & 0.000000   \\
\hline 
\end{tabular}  
\end{adjustbox}
\end{table}

\begin{table}[t]
\caption{Comparison of RBQI vs. Statistical measures and IQA techniques on the S-ReBaQ dataset.}
\label{table:SReBaQEval}
\centering
\scriptsize
\renewcommand{\arraystretch}{1.2}
\begin{tabular}{|ccc|c|c|c|c|c|c|c|}
\hline
\multicolumn{ 2}{|c||}{} & \multicolumn{5}{c|}{\bf{S-ReBaQ}} \\
\hline
 \multicolumn{ 2}{|c||}{} & {\bf PCC} & {\bf SROCC} & {\bf RMSE} &  {\bf P$_{\text{PCC}}$} & {\bf P$_{\text{SROCC}}$} \\
\hline \hline
\parbox[t]{6mm}{\multirow{3}{*}{\rotatebox[origin=c]{90}{\tabular{@{}l}Statistical \\ Measures\endtabular}}} 
& \multicolumn{ 1}{|l||} {\bf AGE} &  0.6453 & 0.6238 & 2.2373 & 0.392900 & 0.000000 \\

&  \multicolumn{ 1}{|l||} {\bf EPs} & 0.4202 & 0.1426 & 1.2049 & 0.000000 & 0.000000 \\

& \multicolumn{ 1}{|l||} {\bf pEPs} &  0.0505 & 0.4990 & 1.6676 & 0.498331 & 0.000000 \\

& \multicolumn{ 1}{|l||} {\bf CEPs} &  0.6283 & 0.6666 & 0.8491& 0.000000 & 0.000000 \\

& \multicolumn{ 1}{|l||} {\bf pCEPs} & 0.8346 & \bf{0.8380} & 0.6011 & 0.000000 & 0.000000 \\
\hline 

\parbox[t]{4mm}{\multirow{3}{*}{\rotatebox[origin=c]{90}{\tabular{@{}l}Image Quality Assessment Metrics \endtabular}}} 
& \multicolumn{ 1}{|l||} {\bf PSNR}   & 0.7099 & 0.6834 & 0.7686 & 0.000000 & 0.000000\\ 

& \multicolumn{ 1}{|l||} {{\bf SSIM}\scriptsize{\cite{SSIM}}} & 0.5975 & 0.5827 & 0.8751 & 0.000000 & 0.000000 \\

& \multicolumn{ 1}{|l||} {{\bf MS-SSIM}\scriptsize{\cite{MSSIM}}} & 0.8048 & 0.8030 & 0.6478 & 0.000000 & 0.000000 \\

& \multicolumn{ 1}{|l||} {{\bf VSNR}\scriptsize{\cite{VSNR}}} & 0.0850 & 0.1717 & 1.0874 & 0.253675 & 0.486686 \\

& \multicolumn{ 1}{|l||}  {{\bf VIF}\scriptsize{\cite{VIF}}} & 0.1027 & 0.2064 & 1.0914 & 0.167842 & 0.005305 \\

& \multicolumn{ 1}{|l||} {{\bf VIFP}\scriptsize{\cite{VIF}}} & 0.6081 & 0.6240 & 0.8664 & 0.000000 & 0.000000 \\

&  \multicolumn{ 1}{|l||} {{\bf UQI}\scriptsize{\cite{UQI}}} & 0.6316 & 0.5932 & 0.8461 & 0.000000 & 0.000000 \\

&  \multicolumn{ 1}{|l||} {{\bf IFC}\scriptsize{\cite{IFC}}} & 0.6235 & 0.6020 & 0.8533 & 0.000000 & 0.000000\\

& \multicolumn{ 1}{|l||} {{\bf NQM}\scriptsize{\cite{NQM}}} & 0.7950 & 0.7816 & 0.6621 & 0.000000 & 0.000000 \\

& \multicolumn{ 1}{|l||} {{\bf WSNR}\scriptsize{\cite{WSNR}}} & 0.7176 & 0.6888 & 0.7601 & 0.000000 & 0.000000 \\

& \multicolumn{ 1}{|l||} {{\bf FSIM}\scriptsize{\cite{FSIM}}} & 0.7243 & 0.7157 & 0.7525 & 0.000000 & 0.000000 \\

& \multicolumn{ 1}{|l||} {{\bf FSIMc}\scriptsize{\cite{FSIM}}} & 0.7278	& 0.7172 & 0.7484 & 0.000000 & 0.000000\\

&\multicolumn{ 1}{|l||}  {{\bf SRSIM}\scriptsize{\cite{SRSIM}}} & 0.7853 & 0.7538 & 0.6757 & 0.000000 & 0.000000 \\

& \multicolumn{ 1}{|l||} {{\bf SalSSIM}\scriptsize{\cite{SalSSIM}}} & 0.7356	 & 0.7300 & 0.7393 & 0.000000 & 0.000000\\

&  \multicolumn{ 1}{|l||} {{\bf CQM}\scriptsize{\cite{CQM}}} & 0.2634 & 0.3645 & 1.0531 & 0.000327 & 0.000276\\

& \multicolumn{ 1}{|l||} {\bf RBQI} & \bf{0.8613} & \bf{0.8222} & \bf{0.5545}  & 0.000000 & 0.000000 \\
\hline 
\end{tabular}  

\end{table}

Tables~\ref{table:ReBaQEval} and~\ref{table:SReBaQEval} show the obtained performance evaluation results of the proposed RBQI technique on the ReBaQ and S-ReBaQ datasets, respectively, as compared to the existing statistical and FR-IQA algorithms. The results show that the proposed quality index yields higher correlation with the subjective scores as compared to any other existing technique. The statistical techniques are shown to not correlate well with the subjective scores on either of the datasets. 
Among the FR-IQA algorithms, the performance of the NQM~\cite{NQM} comes close to the proposed technique for scenes with static background images, i.e., for the ReBaQ$_{\text{static}}$ dataset, as it considers the effects of contrast sensitivity, luminance variations, contrast interaction between spatial frequencies and contrast masking effect while weighting the SNR between the ground truth and reconstructed image. 
The more popular MS-SSIM~\cite{MSSIM} technique is shown to not correlate well with the subjective scores for the ReBaQ database. This is because the MS-SSIM calculates the final quality index of the image by just averaging over the entire image. In the problem of background reconstruction the error might occupy a relatively small area as compared to the image size, thereby under-penalizing the residual foreground. 
RBQI provides a higher correlation with the subjective scores by a margin of 6\% over MS-SSIM on S-ReBaQ.
None of the FR-IQA or statistical techniques were found to correlate with the scores in the ReBaQ$_{\text{dynamic}}$ dataset. This is because the assumption of pixel-to-pixel correspondence is no longer valid in the presence of pseudo-stationary background. The proposed RBQI technique uses a neighborhood window to handle such backgrounds, thereby improving the performance over NQM~\cite{NQM} by a margin of 10\% and by 30\% over MS-SSIM~\cite{MSSIM}. 
CQM~\cite{CQM} used in the Scene Background Modeling Challenge 2016 (SBMC)~\cite{SBMC} and~\cite{BI_Taxonomy} to compare the performance of the algorithms is shown to perform very poorly on all three datasets and hence is not a good choice for evaluating the quality of reconstructed background images and thus is not suitable for comparing the performance of background reconstruction algorithms.

The P-value is the probability of getting a correlation as large as the observed value by random chance, while the variables are independent. If the P-value is less than 0.05 then the correlation is significant. The P-values (P$_{\text{PCC}}$ and P$_{\text{SROCC}}$) reported in Tables~\ref{table:ReBaQEval} and~\ref{table:SReBaQEval} indicate that most of the correlation scores are statistically significant. 


\begin{table}[t]
\centering
\caption{Performance comparison for different values of parameters on the ReBaQ dataset.}
\label{table:simulations} 

\begin{subtable}{\linewidth}\centering 
\caption{Simulation results with different neighborhood search window sizes $nhood$}
\renewcommand{\arraystretch}{1.2}
\scriptsize
\begin{tabular}{|l||r|r|r||r|r|r|}
\hline
\multicolumn{ 1}{|c||}{} &  \multicolumn{3}{c||}{\bf{ReBaQ$_{\text{static}}$}} 
	& \multicolumn{3}{c|}{\bf{ReBaQ$_{\text{dynamic}}$}} \\
\hline
           &  {\bf PCC} & {\bf SROCC} & {\bf RMSE}  &  {\bf PCC} & {\bf SROCC} & {\bf RMSE} \\
\hline
\hline
{\bf nhood=1} & 0.7931 & 0.8314 & 0.5077  & 0.6395 & 0.6539 & 0.5662\\
\hline
{\bf nhood=9} & 0.9015 & 0.8581 & 0.4911 & 0.7834 & 0.6683 & 0.5394\\
\hline
{\bf nhood=17} & 0.9006 & 0.8581 & 0.4837 & 0.7908 & 0.6762 & 0.4374  \\
\hline
{\bf nhood=33} & 0.9001 & 0.8581 & 0.4896 & 0.7818 & 0.6683 & 0.4769 \\
\hline
\end{tabular}  
\label{table:nhoodCor}
\end{subtable}

\vspace*{2mm}
\begin{subtable}{\linewidth}\centering
\caption{Simulation results with different number of scales $L$}
\renewcommand{\arraystretch}{1.2}
\scriptsize
\begin{tabular}{|l||r|r|r||r|r|r|}
\hline
\multicolumn{ 1}{|c||}{} &  \multicolumn{3}{c||}{\bf{ReBaQ$_{\text{static}}$}} 
	& \multicolumn{3}{c|}{\bf{ReBaQ$_{\text{dynamic}}$}} \\
\hline
           &  {\bf PCC} & {\bf SROCC} & {\bf RMSE}  &  {\bf PCC} & {\bf SROCC} & {\bf RMSE} \\
\hline
\hline
{\bf L=1} & 0.8190 & 0.8183 & 0.6667	 & 0.5561 & 0.5520 & 0.7335\\
\hline
{\bf L=2} &0.8597 & 0.8310 & 0.5521 	& 0.7281 & 0.6482 & 0.6050 \\
\hline
{\bf L=3} & 0.9006 & 0.8592 & 0.5077	 & 0.7908 & 0.6773 & 0.5662 \\
\hline
{\bf L=4} & 0.9006 & 0.8581 & 0.4915	 & 0.7954 & 0.6797 & 0.5350 \\
\hline
{\bf L=5} & 0.9006 & 0.8581 & 0.4883	 & 0.8087 & 0.6881 & 0.5191\\
\hline
\end{tabular}  
\label{table:scaleCor}
\end{subtable}

\end{table}


\subsection{Model Parameter Selection}
\label{sec:simulations}
The proposed quality index accepts four parameters:
\begin{inparaenum}
	\item $nhood$, dimensions of the window centered around the current pixel for calculating the $d_s$;
	\item $L$, number of multi-scale levels;
	\item $\beta_s$, used in the calculation of $P_{D,s,l}(x,y)$ in Equation~(\ref{eq:PDsl}); and
	\item $\beta_c$, used in the calculation of $P_{D,c,l}(x,y)$ in Equation~(\ref{eq:PDcl}).
\end{inparaenum}
In Table~\ref{table:simulations}, we evaluate our algorithm with different values for the parameters. These simulations were run only on the ReBaQ dataset. 
Table~\ref{table:simulations}\subref{table:nhoodCor} shows the effect of varying $nhood$ values on the performance of RBQI. The performance of RBQI for ReBaQ$_{\text{static}}$ did not significantly change with the increase in the neighborhood search window size as expected, but the performance of RBQI increased drastically for the ReBaQ$_{\text{dynamic}}$ dataset from $nhood=1$ to $nhood=17$ before starting to drop at $nhood=33$. Thus we chose $nhood=17$ for all our experiments. 
Table~\ref{table:simulations}\subref{table:scaleCor} gives performance results for different number of scales. As a tradeoff between the computation complexity and prediction accuracy we chose the number of scales to be $L=3$. 
The probability summation model parameters $\beta_s$ and $\beta_c$ were found such that they maximized the correlation between RBQI and MOS scores on a training dataset consisting of randomly selected images from the ReBaQ dataset. Values $\beta_s=\beta_c=3.5$ were found to correlate well with the subjective tests. 

These parameters remained unchanged for the experiments conducted on the S-ReBaQ dataset to obtain the values in Table~\ref{table:SReBaQEval}. 

\section{Conclusion}
\label{sec:conclusions}
In this paper we addressed the problem of quality evaluation of reconstructed background images. We first proposed two different datasets for benchmarking the performance of existing and future techniques proposed to evaluate the quality of reconstructed background images. 
Then we proposed the first full-reference Reconstructed Background Quality Index (RBQI) to objectively measure the perceived quality of the reconstructed background images. 

The RBQI uses the probability summation model to combine visual characteristics at multiple scales to quantify the deterioration in the perceived quality of the reconstructed background image due to the presence of any foreground objects or unnaturalness that may be introduced by the background reconstruction algorithm.
The use of a neighborhood search window while calculating the contrast and structure differences provides further boost in the performance in the presence of pseudo-stationary background while not affecting the performance on scenes with static background. 
The probability summation model penalizes only the perceived differences across the reference and reconstructed background images while the unperceived differences do not affect the RBQI, thereby giving better correlation with the subjective scores.
Experimental results on the benchmarking datasets showed that the proposed measure out-performed all the existing statistical and IQA techniques in estimating the perceived quality of reconstructed background images. 


%





\ifCLASSOPTIONcaptionsoff
  \newpage
\fi




\bibliographystyle{myIEEEtran}
\bibliography{IEEEabrv,References}%

\end{document}